\documentclass{article}
\PassOptionsToPackage{numbers,compress}{natbib}

\usepackage[main,final]{neurips_2026}

\usepackage[utf8]{inputenc} 
\usepackage[T1]{fontenc}    
\usepackage{hyperref}       
\usepackage{url}            
\usepackage{booktabs}       
\usepackage{amsfonts}       
\usepackage{nicefrac}       
\usepackage{microtype}      
\usepackage{xcolor}         
\usepackage{graphicx}
\usepackage{subcaption}
\usepackage{amsmath}
\usepackage{amssymb}
\usepackage{algorithm}
\usepackage{algorithmic}
\usepackage{wrapfig}

\title{TENG-BC: Time-Evolving Natural Gradient for High-Accuracy Neural PDE Solvers with General Boundary Conditions}

\author{
  Hongjie Jiang$^{2}$, Di Luo$^{1,3,4}$\thanks{Corresponding author.}
  \\[0.5em]
  $^{1}$State Key Laboratory of Low-Dimensional Quantum Physics and \\
  Department of Physics, Tsinghua University, Beijing, China \\
  $^{2}$Department of Applied Mathematics, Brown University, USA \\
  $^{3}$Frontier Science Center for Quantum Information, Beijing, China \\
  $^{4}$C.N. Yang Institute of Advanced Study, Tsinghua University, Beijing, China
  \\[0.3em]
  \texttt{hongjie\_jiang@brown.edu},
  \texttt{diluo@tsinghua.edu.cn}
}

\begin{document}

\maketitle
\begin{abstract}
  Accurately solving time-dependent partial differential equations (PDEs) with neural networks remains challenging due to long-time error accumulation and the difficulty of enforcing general boundary conditions. We introduce TENG-BC, a high-precision neural PDE solver based on the Time-Evolving Natural Gradient, designed to perform under generic boundary constraints. At each time step, TENG-BC performs a boundary-aware optimization that jointly enforces interior dynamics and boundary conditions, accommodating Dirichlet, Neumann, Robin, and mixed types within a unified framework. This formulation admits a natural-gradient interpretation, enabling stable time evolution without delicate penalty tuning. Across benchmarks over diffusion, transport, and nonlinear PDEs with various boundary conditions, TENG-BC achieves solver-level accuracy under comparable sampling budgets, outperforming conventional solvers and physics-informed neural network (PINN) baselines.
\end{abstract}

\section{Introduction}

Accurately and efficiently solving time-dependent partial differential equations (PDEs) remains a cornerstone challenge in computational science.
Such equations govern a wide range of physical and engineering systems, from heat conduction and fluid transport to wave propagation, quantum dynamics, and reactive processes.
Classical numerical solvers based on finite-difference, finite-element, and spectral methods achieve high accuracy through carefully designed discretizations and operator-specific schemes.
However, their performance is tightly coupled to mesh quality, stencil design, and boundary handling.

\begin{figure*}[h]
    \centering
    \vspace{-5pt}
    \includegraphics[width=0.9\linewidth,height=0.39\linewidth]{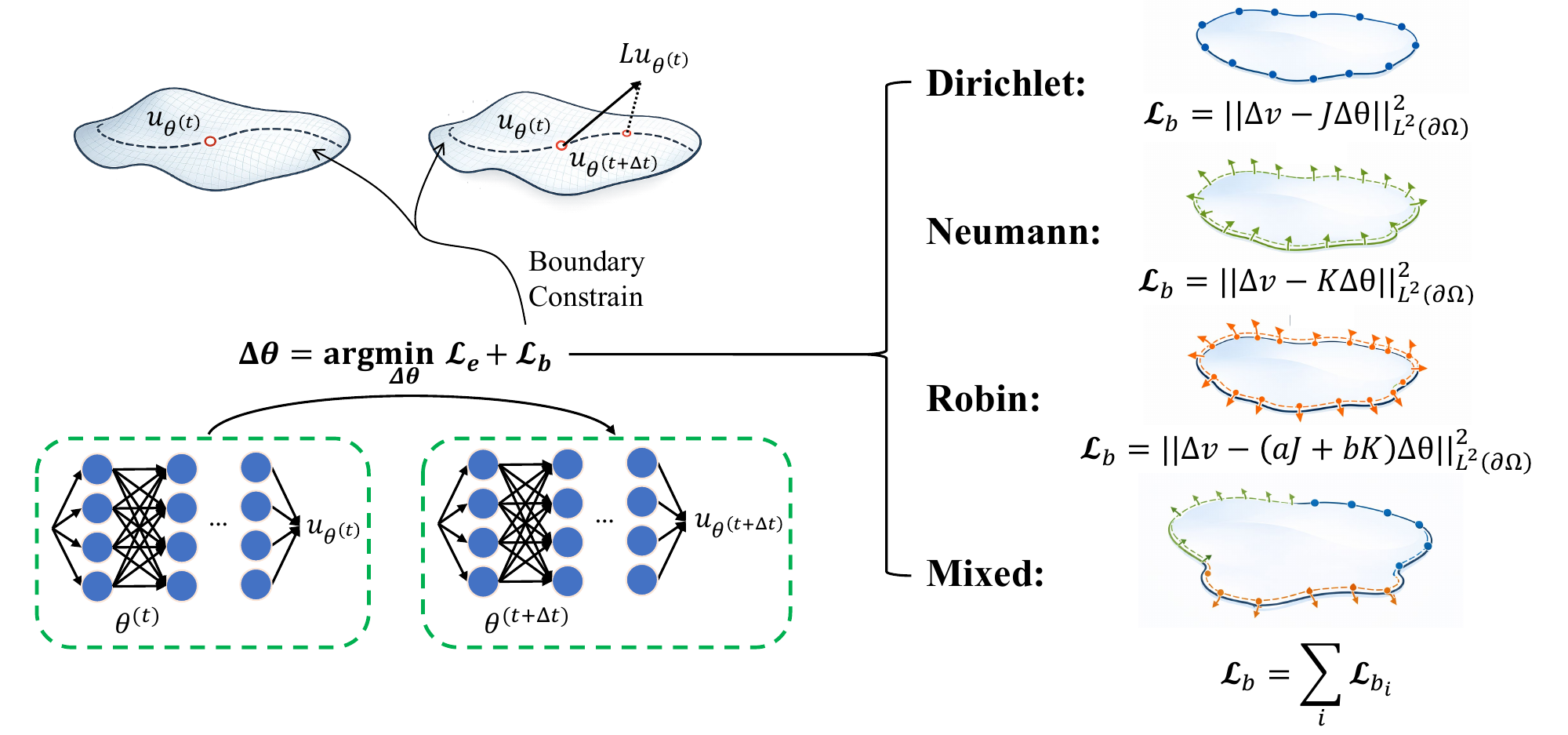}
    \caption{
    \textbf{Overview of the boundary-aware TENG-BC update.}
    The neural representation is advanced by solving a unified least-squares problem that enforces both interior dynamics and boundary constraints.
    Different boundary types are incorporated through the same optimization operator, which is repeatedly applied to realize temporal evolution. }
    \label{fig1:flowchart}
    \vspace{-5pt}
\end{figure*}

Recent years have witnessed rapid progress in neural-network-based PDE solvers
\cite{han2017deep, yu2018deep, long2018pde, carleo2017solving, raissi2019physics, li2020fourier, lu2019deeponet, sirignano2018dgm, weinan2021algorithms, chen2022simulating, chen_2023},
which represent $u(x,t)$ by a neural network representation $u_\theta(x,t)$ and bypass explicit meshes.
Despite their flexibility, most existing neural solvers face two recurring obstacles in time-dependent settings.
(\emph{i}) Global-in-time training: many approaches minimize a single space--time residual over $[0,T]$
\cite{raissi2019physics, wang2023long, wang2021learning, wang2021spacetime},
which couples all time steps into one objective and provides weak control over local temporal errors; small global loss may still permit stepwise inaccuracies that propagate and amplify over time
\cite{krishnapriyan2021characterizing, wang2024respecting}.
(\emph{ii}) Boundary enforcement: physical boundary conditions are commonly handled by soft penalty terms\cite{raissi2019physics} that require careful weighting and often under-enforce Neumann, Robin or mixed boundaries. Alternatively, imposing hard boundary constraints that construct boundary-satisfying trial spaces
\cite{lu2021physics, Sukumar_2022} can be brittle for complex geometries, heterogeneous boundary segments, or time-dependent boundary data.
These two issues compound in long-time simulations, where global objectives and imperfect boundary handling can jointly induce instability and reduce accuracy.

A promising remedy to the first issue is provided by the recently proposed \emph{Time-Evolving Natural Gradient} (TENG) framework \cite{chen2024teng},
which replaces global-in-time optimization by a sequence of localized, per-step functional projections.
TENG achieves solver-level accuracy and strong long-time stability by controlling temporal evolution step by step.
However, existing TENG formulations primarily focus on periodic or simple boundary settings.

\paragraph{Contributions.}
In this work, we present \textbf{TENG-BC}, a boundary-aware formulation of time-evolving neural PDE solvers that achieves high accuracy under generic boundary conditions.
The main contributions of this work are:

\begin{itemize}
    \item \textbf{Generic boundary-aware TENG-BC Framework.} We introduce a unified boundary-aware least-squares formulation (see Figure~\ref{fig1:flowchart}) for time-evolving neural PDE solvers, enabling the consistent treatment of Dirichlet, Neumann, Robin, and mixed boundary conditions within a single optimization operator.
    \item \textbf{Scalable boundary-aware least-squares algorithm.} We develop an efficient unified implementation that assembles interior and boundary constraints into a single linearized least-squares system, enabling consistent, linearly scalable updates across boundary types.
    \item \textbf{High-precision accuracy across diverse PDE regimes.} TENG-BC delivers stable, high-precision solutions across diffusion-dominated, advection-dominated, and nonlinear regimes, consistently outperforming both classical numerical solvers and PINN-based baselines.
\end{itemize}

\section{Related Work}
\subsection{Neural solvers for PDEs}

Neural networks have been widely adopted as function approximators for partial differential equations, enabling solution representations that are independent of predefined spatial meshes.
Many existing equation-driven approaches treat time as an additional input variable and optimize the neural representation over the entire space--time domain.
This strategy underlies Physics-Informed Neural Networks (PINNs) \cite{raissi2019physics, wang2023long, wang2021learning, sirignano2018dgm}.
An alternative line of work advances neural solutions through time by updating the representation sequentially.
In this setting, the neural network is interpreted as a time-dependent ansatz, and evolution is realized through repeated local updates rather than global regression.
Representative examples include methods derived from the time-dependent variational principle \cite{dirac1930note, koch2007dynamical, carleo2017solving, Du_2021, berman2023randomized}, which project the governing dynamics onto a parameterized manifold, as well as optimization-based time integration schemes \cite{kochkov2018variational, chen2023implicit, gutierrez2022real, luo2022autoregressive, luo2023gauge, Sinibaldi2023unbiasingtime}.
Beyond equation-driven solvers, neural networks have also been employed to learn PDE dynamics directly from data.
This includes operator-learning frameworks such as the Fourier Neural Operator \cite{li2020fourier} and DeepONet \cite{lu2019deeponet}.
Rather than approximating a single solution trajectory, these models aim to capture mappings between function spaces and are often trained across families of PDEs.

\subsection{Boundary treatment in neural PDE solvers}

In neural PDE solvers, boundary constraints are typically incorporated through either soft or hard enforcement mechanisms. 
A widely used strategy is to impose boundary conditions through penalty terms added to the training objective.
This approach is adopted by most PINN-based methods, where boundary residuals are included alongside interior PDE residuals with manually chosen weights \cite{raissi2019physics, wang2021learning}.
An alternative line of work seeks to enforce boundary conditions through hard constraints by construction.
This is commonly achieved by modifying the neural representation to satisfy boundary conditions identically, for example through tailored trial spaces and distance functions \cite{lu2021physics, Sukumar_2022}.
Beyond these two paradigms, several other structure-aware strategies include \cite{weinan2021algorithms, liu2023domainagnosticfourierneural} have also been explored.

More recently, various techniques have been developed to incorporate physical constraints into neural models beyond the conventional treatment of boundary residuals.
These include probabilistic corrections, modifications of neural operators, and constraint-enforcing procedures based on projection or optimization \cite{hansen2023probconserv, saad2023boon, cheng2025eci, utkarsh2026probharde2e}.
Such methods provide alternative ways to improve physical consistency without relying solely on penalty-based training objectives.
Unlike these approaches, TENG-BC incorporates boundary operators directly into the local linearized least-squares projection at each time step, extending the sequential neural PDE evolution framework of TENG to general boundary conditions.
In this way, boundary constraints are taken into account when determining the parameter updates, while the natural-gradient projection structure of TENG is preserved.

\section{Methods}
\subsection{Problem Formulation}

We consider time-dependent partial differential equations (PDEs) of the form
\begin{equation*}
u_t(x,t) = Lu(x,t), \quad x\in\Omega,~t\in[0,T],
\end{equation*}
subject to general mixed boundary conditions imposed on disjoint subsets of $\partial\Omega$,
\begin{align*}
[a(x)u(x,t) + b(x)\partial_n u(x,t)]|_{\partial\Omega} = v(x,t),
\end{align*}
where $L$ is a spatial differential operator and $\partial_n$ denotes the normal derivative on the boundary.
When representing mixed conditions, different coefficient values are taken on different segments of the boundary. 

Contrast with global-in-time methods, TENG represents the temporal evolution of $u(x,t)$ as a sequence of parameter-space updates on a time-dependent representation $u_{\theta^{(t)}}(x)$.
Here $\theta^{(t)}$ denotes the set of parameters associated with the solution at the time step $t$.
TENG performs local functional optimization at each time step to update the parameters $\theta^{(t)} \to \theta^{(t+\Delta t)}$ in accordance with the governing PDE
\begin{equation*}
\theta^{(t+\Delta t)} = \arg\min_\theta \big(
\mathcal{L}_e(u_\theta) +
\mathcal{L}_b(u_\theta)
\big),
\end{equation*}
where $\mathcal{L}_e$ and $\mathcal{L}_b$ denote the interior and boundary consistency terms.
Specifically, $\mathcal{L}_e(u_\theta):=\frac{1}{2}||u_\theta - u_\text{target}||_{L^2(\Omega)}^2$ enforces the temporal update $u_t=Lu$, while $\mathcal{L}_b(u_\theta):=\frac{1}{2}||a u_\theta + b \partial_n u_\theta - v||_{L^2(\partial\Omega)}^2$ encodes boundary constraints.

Ignoring boundary effects momentarily, the update at a fixed time step is obtained by solving a local least-squares problem in parameter space.
Given parameters $\theta$, we seek an increment $\Delta\theta$ such that the first-order variation of the neural function best matches the target field $u_{\text{target}}$ in the $L^2(\Omega)$ sense:
\begin{equation*}
    \Delta\theta_e = \arg\min_{\Delta\theta}\left|\left|\Delta u - J\Delta\theta\right|\right|_{L^2(\Omega)}^2,
\end{equation*}
where $J=\partial_\theta u_\theta$ and $\Delta u = u_\text{target}-u_\theta$. It has been proved in \cite{chen2024teng} that this update corresponds to the natural-gradient step \cite{muller2023achieving} in function space, yielding a quasi-Newton update that projects the continuous PDE evolution onto the neural parameter manifold.

\subsection{Challenges of Penalty-Based Boundary Enforcement}

A natural way to incorporate boundary conditions is to penalize the boundary mismatch
\[
\|\mathcal{B}[u_\theta]-v\|_{L^2(\partial\Omega)}^2,
\quad
\mathcal{B}u = a u + b\partial_n u .
\]
While this strategy is standard in global-loss formulations such as PINNs, it is not sufficient for the time-stepping mechanism of TENG.
TENG does not evolve the solution by minimizing a residual loss over the current neural function; instead, each step computes a parameter increment \(\Delta\theta\) by matching the neural output \(u_\theta\) to a target field.

For general boundary conditions, however, the prescribed data specify the value of the boundary operator
\[
\mathcal{B}u = a u + b\partial_n u,
\]
rather than a target value of \(u\) itself.
Because \(\partial_n\) is a differential operator, such operator-valued boundary data cannot, in general, be converted into an equivalent pointwise target function for \(u_\theta\).
Thus, the original TENG geometry based only on fitting \(u_\theta+J\Delta\theta\) is insufficient for general boundary constraints.

The key challenge is therefore not to add another penalty term, but to extend the local projection geometry so that it also contains boundary-operator variations.
In particular, the update space must include the parameter sensitivity of the normal derivative, through
\[
K=\partial_\theta(\partial_n u_\theta).
\]
By incorporating boundary-operator sensitivities into the local projection, TENG-BC extends the original TENG update to account for general boundary conditions while preserving its sequential time-stepping structure.

\subsection{TENG-BC: TENG with arbitrary boundaries}

\paragraph{Boundary-aware optimization.}
Motivated by the above challenge, TENG-BC augments the local TENG projection geometry by incorporating boundary-operator variations. General mixed boundary conditions lead to the following least-squares problem
\begin{equation}
\label{eq:boundary_ls}
\Delta\theta_b = \arg\min_{\Delta\theta} \|\Delta v - (aJ + bK)\Delta\theta\|_{L^2(\partial\Omega)}^2,
\end{equation}
where $J=\partial_\theta u_\theta$, $K=\partial_\theta(\partial_n u_\theta)$, and
$\Delta v = v - (a u_\theta + b\partial_n u_\theta)$ denotes the instantaneous boundary mismatch.
The solution of Equation~\ref{eq:boundary_ls} represents the parameter variation required to enforce boundary consistency at the current time step.
The detailed derivation of this boundary-aware least-squares formulation is provided in Appendix~\ref{appendix-a: ls drivation}.

The unified formulation in Equation.\ref{eq:boundary_ls} recovers standard boundary conditions as special cases:

\emph{Dirichlet boundary.}
For $u(x,t)=v(x,t)$, we set $(a,b)=(1,0)$, yielding
\[
\Delta v = v - u_\theta, \quad \text{with operator } J.
\]

\emph{Neumann boundary.}
For $\partial_n u(x,t)=v(x,t)$, we set $(a,b)=(0,1)$, yielding
\[
\Delta v = v - \partial_n u_\theta, \quad \text{with operator } K.
\]

\emph{Robin boundary.}
For $a u + b \partial_n u = v$ with $a,b\neq 0$, we obtain
\[
\Delta v = v - (a u_\theta + b \partial_n u_\theta), \quad \text{with operator } aJ + bK.
\]

\emph{Mixed boundary conditions.}
In the general case, $(a(x), b(x))$ varies spatially along $\partial\Omega$, so that different boundary types are specified on different segments.
All such constraints are assembled within the same operator and enforced simultaneously through a single least-squares update.

\paragraph{Combined interior--boundary update.}
The actual TENG-BC step at each time increment is obtained by combining interior consistency and boundary constraints within a single local least-squares problem.
Specifically, the parameter update $\Delta\theta$ is computed as
\begin{equation}
\label{eq:full_ls}
\begin{aligned}
\Delta\theta = &\arg\min_{\Delta\theta} \Big( \|\Delta u - J\Delta\theta\|_{L^2(\Omega)}^2 + \\
&\|\Delta v - (aJ + bK)\Delta\theta\|_{L^2(\partial\Omega)}^2 \Big).
\end{aligned}
\end{equation}
This formulation can be equivalently viewed as a unified least-squares problem over the domain closure $\overline{\Omega}$, where interior and boundary residuals are assembled under the combined $L^2(\overline{\Omega})$ inner product. 
From a geometric viewpoint, this unified least-squares update also admits an interpretation as a natural-gradient step in function space.
A detailed derivation and analysis are provided in Appendix~\ref{appendix-b: natgrad}.

Its practical implementation at each time step is summarized in Algorithm~\ref{alg:teng_boundary}. 
At each update, the parameter increment is obtained by explicitly assembling an augmented design matrix $\mathcal{J}$ that stacks interior and boundary Jacobian blocks, and solving the resulting linearized least-squares system, rather than backpropagating through the full boundary loss.
Because the Jacobian contributions from interior samples and boundary samples are combined into one unified matrix, all boundary types are handled within the same computational pipeline without introducing specialized branches.
As a result, the per-step computational cost scales linearly with the number of sampled interior and boundary points.

Compared with existing boundary-handling strategies, the proposed formulation has two key advantages.
It does not rely on boundary-specific trial-space constructions, and thus applies uniformly to general and time-dependent boundary conditions.
Instead, boundary constraints are incorporated directly into the update operator, rather than enforced via penalty-based weighting.

\begin{algorithm}[t]
\caption{Boundary-Aware TENG Update at One Time Step}
\label{alg:teng_boundary}
\begin{algorithmic}[1]
\STATE \textbf{Input:} Current parameters $\theta^{(t)}$, target field $u_{\text{target}}$, boundary data $(a,b,v)$
\STATE \textbf{Output:} Updated parameters $\theta^{(t+\Delta t)}$
\vspace{0.5em}

\STATE Initialize $\theta \leftarrow \theta^{(t)}$
\FOR{$k = 1$ to $N_{\mathrm{LS}}$}
    \STATE Compute mismatch $\Delta u = u_{\text{target}} - u_\theta$ on $\Omega$, and $\Delta v = v - (a u_\theta + b \partial_n u_\theta)$ on $\partial\Omega$
    \STATE Compute Jacobians $J_\Omega = \partial_\theta u_\theta$ on $\Omega$, and $J_{\partial\Omega} = \partial_\theta u_\theta$, $K_{\partial\Omega} = \partial_\theta(\partial_n u_\theta)$ on $\partial\Omega$
    \STATE Form the unified residual $\mathcal{R} = (\Delta u, \Delta v)$
    \STATE Form the unified Jacobian $\mathcal{J} = (J_\Omega,\ aJ_{\partial\Omega} + bK_{\partial\Omega})$
    \STATE Solve the unified least-squares problem
    $
    \Delta\theta \gets
    \arg\min_{\Delta\theta}
    \|\mathcal{R} - \mathcal{J}\Delta\theta\|_{L^2(\overline{\Omega})}^2
    $
    \STATE Update parameters $\theta \leftarrow \theta + \Delta\theta$
\ENDFOR
\STATE Set $\theta^{(t+\Delta t)} \leftarrow \theta$
\end{algorithmic}
\end{algorithm}

\paragraph{Temporal discretization.}
Different time integration schemes correspond to different constructions of the target function $u_{\text{target}}$.
In the simplest case, the \textbf{TENG--Euler} scheme uses
\begin{equation*}
u_{\text{target}} = u_{\theta^{(t)}} + \Delta t\, L(u_{\theta^{(t)}}).
\end{equation*}
Higher-order schemes, such as \textbf{TENG--Heun} and \textbf{TENG--RK4} (see detailed time evolving schemes in Appendix~\ref{appendix-c: settings}), introduce multiple intermediate evaluations of the operator $L(\cdot)$ to form more accurate targets, while keeping the same optimization mechanism at each substage.
Importantly, the least-squares stepper used to fit each target remains identical across different integration orders.

\subsection{Complexity and Error Analysis}
To characterize the behavior of the proposed time-stepping scheme, we analyze its computational complexity and error propagation.
Denote by \(N\) the number of sampling points and by \(P_{\text{eff}}\) the number of parameters involved in the update.
The dominant cost arises from forming and solving the least-squares problems, leading to a complexity of approximately $\mathcal{O}(N P_{\text{eff}}^2)$.
In practice, we adopt a partial-parameter update strategy with \(P_{\text{eff}} \ll P\) to reduce the computational cost while maintaining accuracy.

Computing the normal-derivative Jacobian $K$ introduces additional automatic differentiation overhead compared with the standard output Jacobian $J$.
However, this computation is required only at boundary sampling points, which are substantially fewer than interior points in our experiments.
For the heat equation experiments in Section~\ref{sec:heat}, the per-step runtime is 4.26 seconds with Dirichlet boundary conditions and 4.29 seconds with Neumann boundary conditions, suggesting that the additional cost of computing $K$ is modest in this setting.

We also briefly characterize the resulting error propagation. The error satisfies $\epsilon(t)\;\lesssim\;\sum_n G^n (\tau_n + r_n)$,
where $\tau_n$ and $r_n$ denote the time discretization and projection errors at each step.
Under Euler time evolution, $\tau_n = \mathcal{O}(\Delta t)$ (and higher-order for Heun/RK4), while the projection error accumulates through $r_n$ depending on the stability factor $G$.
This highlights that accuracy is governed by the least-squares residual at each step, which is directly controlled in the proposed update. Details are provided in Appendix~\ref{appendix-c: error_bound}.

\subsection{Implementation Details}

The practical implementation of TENG-BC involves several design considerations that ensure stable and accurate optimization in high-dimensional neural representations. 

\emph{(i) Spatial Sampling and Discrete Least-Squares Approximation.} 
Interior and boundary points are sampled on a fixed grid and remain unchanged across time steps.
The continuous $L^2$ objective is approximated via discrete least-squares using Monte Carlo quadrature.

\emph{(ii) Numerical Stabilization and Singular-Value Truncation.}
Least-squares updates are computed using singular value decomposition (SVD) in double precision, with a truncation threshold of $10^{-7}$ to suppress near-singular directions and improve numerical stability.

\emph{(iii) Partial Parameter Updates for Stable Optimization.}
At each step, only a subset of parameters is optimized, reducing computational cost and improving conditioning.

\emph{(iv) Robustness to Random Parameter Selection.}
Random active-parameter selection has limited impact on numerical accuracy, with RMSE variations on the order of $10^{-7}$ across different random seeds, demonstrating the reproducibility of the results.

Together, these implementation strategies ensure that the TENG-BC framework achieves both stability and precision across diverse PDE systems.  
They bridge the gap between the continuous functional formulation and its discrete neural realization, enabling robust training dynamics and high-fidelity solution accuracy in all experiments presented below.

\section{Results}

\subsection{Experimental Setup}

To evaluate the capability and robustness of TENG-BC, we conduct a series of benchmark experiments spanning representative types of PDE dynamics.  
These include diffusion-dominated (\textit{heat equation}), advection-dominated (\textit{transport equation}), and nonlinear coupled regimes (\textit{viscous Burgers equation}), each designed to highlight distinct numerical and physical challenges such as stability, sharp-front preservation, and multiple boundary conditions.  

In all cases, the neural representation $u_{\theta^{(t)}}(x)$ is modeled by a fully connected network (see Appendix~\ref{appendix-c: settings}) whose parameters are updated through the TENG-BC algorithm described above. 
The network takes the spatial coordinates $x \in \Omega \subset \mathbb{R}^d$ as input and outputs a scalar field value.
All experiments share a common optimization pipeline to ensure cross-equation comparability.
In all experiments, TENG-BC advances the solution using a fixed time step size $\Delta t = 5\times10^{-4}$.

For comparison, we include representative baselines from both neural and conventional paradigms.
On the neural network side, we consider a standard Physics-Informed Neural Network (PINN) trained with BFGS or ENGD optimizer \cite{muller2023achieving}, as well as the recent CoPINN method \cite{duancopinn}, which introduces cognitive sampling to improve training efficiency.
On the conventional method side, we include a finite element method (FEM) baseline \cite{skfem2020} using quadratic (Q2) elements. The FEM resolution denotes the number of elements along each spatial direction; for example, FEM-256 corresponds to a $256\times256$ mesh with $513^2$ scalar degrees of freedom.
Each equation is solved by TENG-BC with identical network architecture and comparable sampling density to ensure fairness across tasks.
Further implementation details (data sampling strategy, numerical precision, and optimization hyperparameters) are provided in Appendix~\ref{appendix-c: settings}, together with the full specifications of each PDE system, including initial and boundary conditions, baseline solver configurations, and all parameter settings for reproducibility.

\subsection{Heat Equation with Different Boundaries}
\label{sec:heat}
\begin{figure}[t]
    \centering
    \begin{subfigure}{0.8\linewidth}
        \centering
        \includegraphics[width=\linewidth,height=0.5\linewidth]{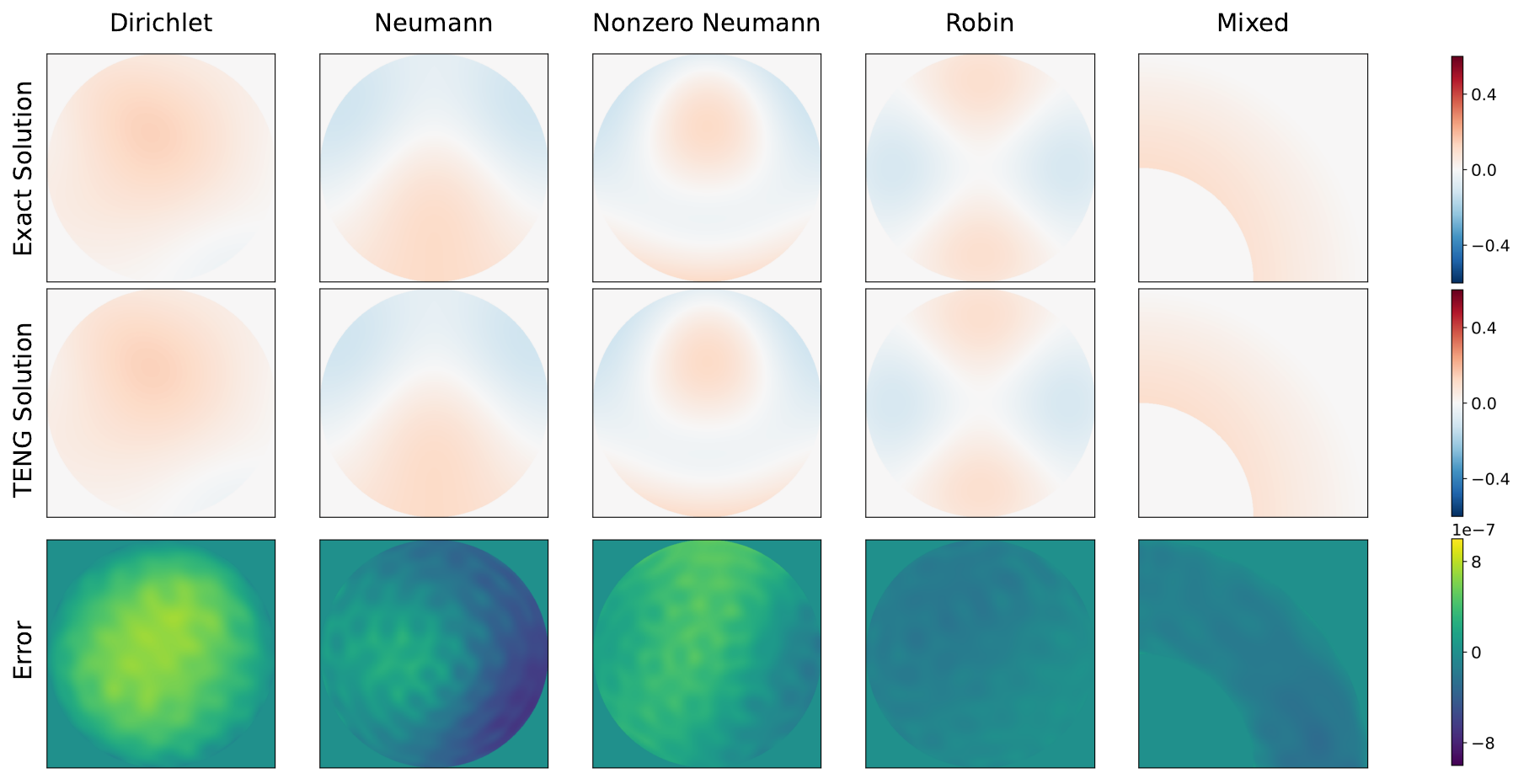}
        \caption{}
        \label{fig2-a:sub1}
    \end{subfigure}
    \begin{subfigure}{0.95\linewidth}
        \centering
        \includegraphics[width=\linewidth,height=0.22\linewidth]{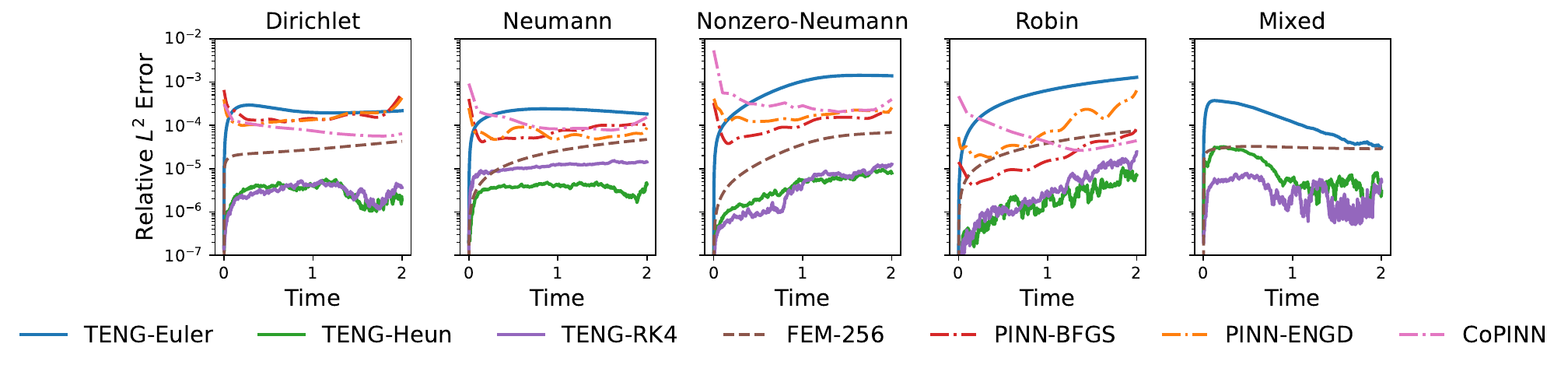}
        \caption{}
        \label{fig2-c:sub3}
    \end{subfigure}
    \caption{
    \textbf{Performance of TENG-BC on heat equation with different boundary conditions.}
    \textbf{(a)} TENG-Heun predictions at $T=1.0$ are shown with analytic references and pointwise error maps.
    \textbf{(b)} Per-time-step relative $L^2$-error comparison among TENG-BC variants and baselines.
    }
    \label{fig2:heat_loss}
\end{figure}

We first evaluate the proposed TENG-BC framework on the two-dimensional heat equation,
\begin{equation*}
    u_t = \kappa\nabla^2u
\end{equation*}
to examine its capability of handling different boundary formulations.
Four boundary types are considered: Dirichlet, Neumann, Mixed, and Robin.
For the Neumann case, both zero and nonzero flux conditions are tested to assess sensitivity to boundary value.
All simulations are conducted on the circular domain $\{r\leq 1\}$, except for the mixed-boundary condition, which is computed on a quarter-annulus region $\{0.5\leq r\leq1,\,0\leq\theta\leq\frac{\pi}{2}\}$.
For quantitative evaluation, the analytic solution is used as the reference field.

Figure~\ref{fig2:heat_loss} summarizes the quantitative performance of TENG-BC, illustrating how different algorithmic variants behave under identical experimental settings.
To assess the influence of temporal propagation on overall accuracy, we test three time discretization schemes of different formal orders.
As shown in Table~\ref{tab:heat}, both TENG-Heun and TENG-RK4 achieve comparable accuracy and significantly outperform the first-order TENG-Euler across most boundary conditions, consistent with their reduced local truncation error.
The Heun variant offers an attractive balance between precision and computational efficiency, reaching nearly the same accuracy as RK4 while requiring roughly half the training cost.
This suggests that the projection error may become the dominant source of error for higher-order schemes, limiting the additional accuracy gains from RK4 over Heun.

\begin{table}[t]
\centering
\caption{
    \textbf{Full-time-range relative $L^2$-error of TENG-BC with three propagation schemes on heat equation with different boundary conditions.} 
    }\label{tab:heat}%
\vspace{5pt}
\begin{tabular}{@{}cccccc@{}}
\toprule
$L^2$-error$\downarrow$ & Dirichlet & Neumann & Nonzero Neumann & Robin & Mixed \\
\midrule
TENG-Euler & 2.17e-04 & 2.05e-04 & 9.04e-04 & 6.40e-04 & 1.25e-04 \\
TENG-Heun & 3.08e-06 & \textbf{3.60e-06} & \textbf{4.41e-06} & \textbf{2.42e-06} & 1.15e-05 \\
TENG-RK4 & \textbf{3.07e-06} & 1.14e-05 & 4.72e-06 & 5.00e-06 & \textbf{3.31e-06} \\
\bottomrule
\end{tabular}
\end{table}

Figure~\ref{fig2-a:sub1} visualizes the predicted solutions by TENG-Heun and their deviations from the analytic reference at $T=1$. 
Across all boundary types, TENG-BC accurately captures the reference profiles with small localized errors, including the mixed-boundary case where distinct constraints coexist on different boundary segments.
The accumulated spatial error remains at the $10^{-7}$ level, confirming that the per-step least-squares fitting effectively preserves both physical smoothness and long-term stability.

Figure~\ref{fig2-c:sub3} compares the per-time-step relative $L^2$ errors of TENG-BC variants and baselines.
Across the entire evolution range, TENG-Heun and TENG-RK4 maintain markedly lower relative error and smoother temporal variation than the baseline methods.
The PINN curves remain relatively large throughout the time span, indicating difficulty in accurate convergence and temporal dependency capture.
FEM starts with smaller initial errors due to its mesh-based representation of the initial condition, but deteriorates over time as discretization and interpolation errors accumulate.
In contrast, TENG-BC sustains uniformly low error with minimal long-time drift, showing that the local least-squares updates preserve temporal consistency.

To assess the contribution of boundary-aware projection, we compare TENG-BC with the original TENG and a soft-penalty baseline that enforces Dirichlet boundary conditions through penalty-based optimization.
As shown in Table~\ref{tab:heat_ablation}, the original TENG exhibits substantial errors without explicit boundary enforcement.
Introducing a soft boundary penalty considerably improves both interior and boundary accuracy, but the resulting errors remain significantly larger than those of TENG-BC.
These results highlight the effectiveness of directly incorporating boundary constraints into the local least-squares projection, rather than relying on conventional penalty-based optimization.

\begin{table}[t]
\centering
\caption{
\textbf{Comparison of boundary enforcement strategies for the heat equation with Dirichlet boundary conditions.}
RMSE and bRMSE denote the errors in the domain and on the boundary, respectively.
}
\label{tab:heat_ablation}
\vspace{5pt}
\begin{tabular}{@{}lcccc@{}}
\toprule
Method & \multicolumn{2}{c}{$T=1.0$} & \multicolumn{2}{c}{$T=2.0$} \\
\cmidrule(lr){2-3}\cmidrule(l){4-5}
& RMSE$\downarrow$ & bRMSE$\downarrow$ & RMSE$\downarrow$ & bRMSE$\downarrow$ \\
\midrule
TENG-BC & \textbf{5.92e-06} & \textbf{4.76e-08} & \textbf{7.54e-06} & \textbf{2.81e-08} \\
TENG-Soft Penalty & 1.77e-03 & 6.77e-04 & 1.19e-03 & 8.30e-04 \\
TENG & 8.11e-01 & 3.11e-01 & 2.96e+00 & 2.07e+00 \\
\bottomrule
\end{tabular}
\vspace{-5pt}
\end{table}

\subsection{Transport Equation on Circular Domain}

\begin{wrapfigure}[13]{r}{0.46\textwidth}
    \vspace{-24pt}
    \centering
    \includegraphics[width=\linewidth]{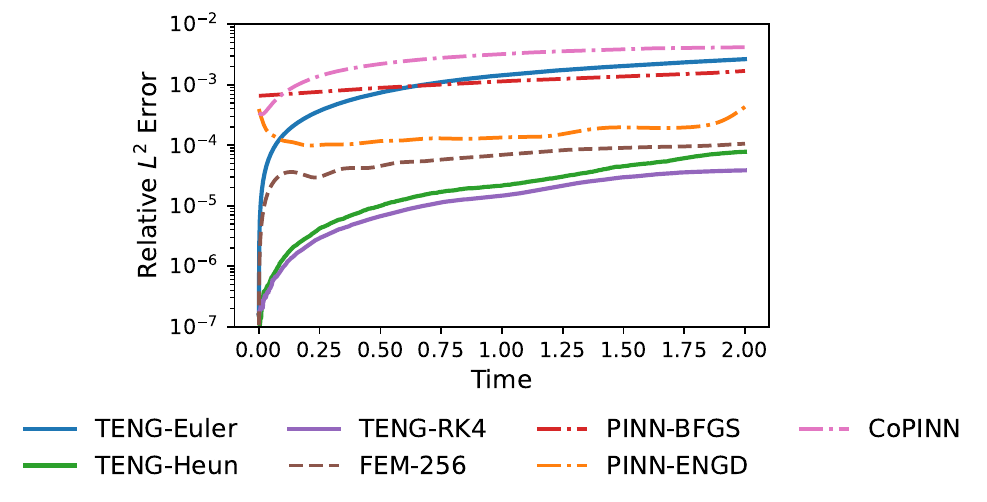}
    \vspace{-8pt}
    \caption{
    \textbf{Performance of TENG-BC on the transport equation with nonzero Dirichlet boundary.}
    Per-time-step relative $L^2$ errors of TENG-BC and baselines.
    }
    \label{fig3:transport}
    \vspace{-10pt}
\end{wrapfigure}

We further evaluate TENG-BC on the two-dimensional transport equation
\begin{equation*}
    \partial_t u + v(x)\cdot \nabla u = 0
\end{equation*}
which governs purely advective dynamics.
Unlike the heat equation, where diffusion naturally damps perturbations and dissipates error, the transport equation propagates information along characteristic directions, causing any local mismatch to persist and gradually accumulate over time.
This setting therefore provides a stringent test of numerical stability and accuracy for function-based solvers.

The simulation is performed on the circular domain $\{r \le 1\}$ with an inhomogeneous Dirichlet boundary that injects a spatially varying inflow profile, preventing convergence to a trivial steady state and emphasizing non-equilibrium transport.
The analytic reference is constructed by tracing the initial and boundary conditions along characteristic trajectories.

As reported in Figure~\ref{fig3:transport}, TENG-BC maintains consistently lower error than PINN, CoPINN, and FEM baselines, demonstrating stable propagation under purely advective dynamics.
Notably, this level of accuracy is achieved despite the absence of dissipative mechanisms, under which error growth is typically expected.

\subsection{Burgers Equation with Dirichlet Boundary}

We finally evaluate TENG-BC on the two-dimensional viscous Burgers equation,
\begin{equation*}
    u_t + u(\partial_{x_1}u+\partial_{x_2}u) = \nu\nabla^2u,
\end{equation*}
with Dirichlet boundary and a small diffusion coefficient $\nu = 10^{-3}$.
This configuration produces highly nonlinear dynamics with sharp gradients and near-discontinuous transitions in the solution profile, a regime notoriously difficult for both neural and conventional numerical solvers.
Since the viscous Burgers equation lacks a closed-form analytic solution, a spectral solver \cite{canuto2007spectral} with $1024$ modes is employed as the reference field.

\begin{figure}
    \centering
    \begin{subfigure}{0.63\linewidth}
        \centering
        \includegraphics[width=\linewidth,height=0.55\linewidth]{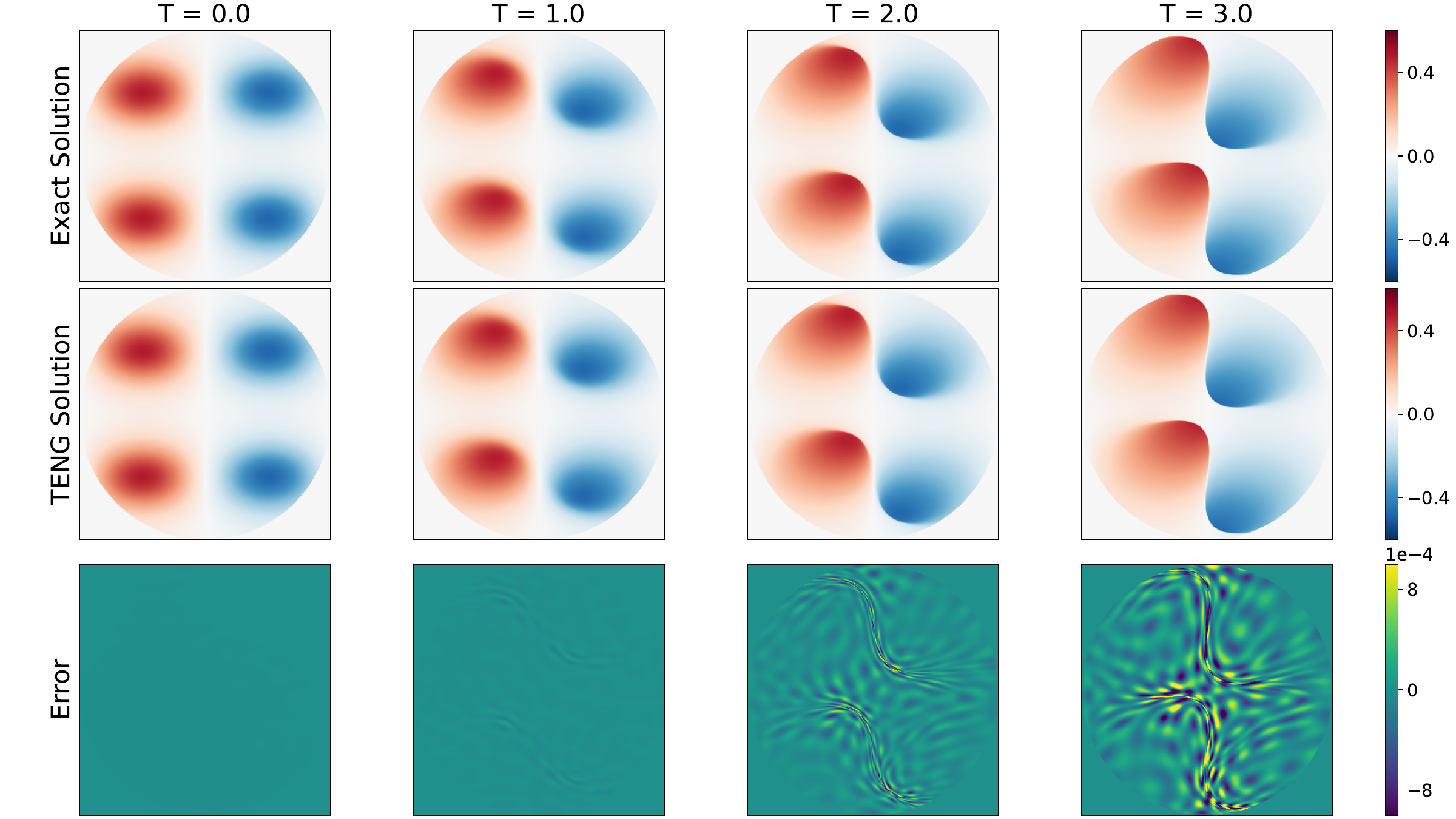}
        \caption{}
        \label{fig4-a:sub1}
    \end{subfigure}
    \begin{subfigure}{0.38\linewidth}
        \centering
        \includegraphics[width=\linewidth]{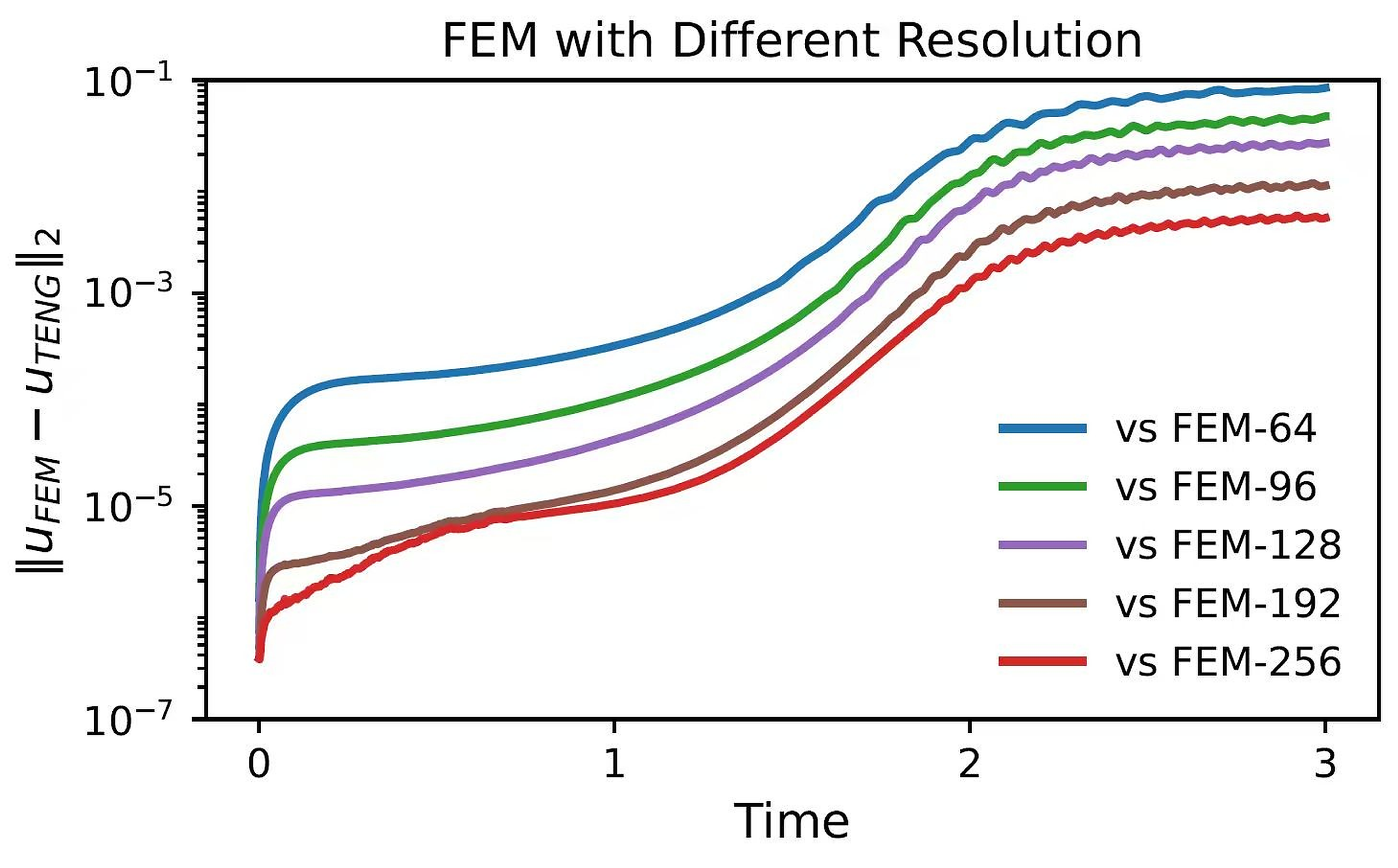}
        \caption{}
        \label{fig4-b:sub2}
    \end{subfigure}
    \begin{subfigure}{0.23\linewidth}
        \centering
        \includegraphics[width=\linewidth]{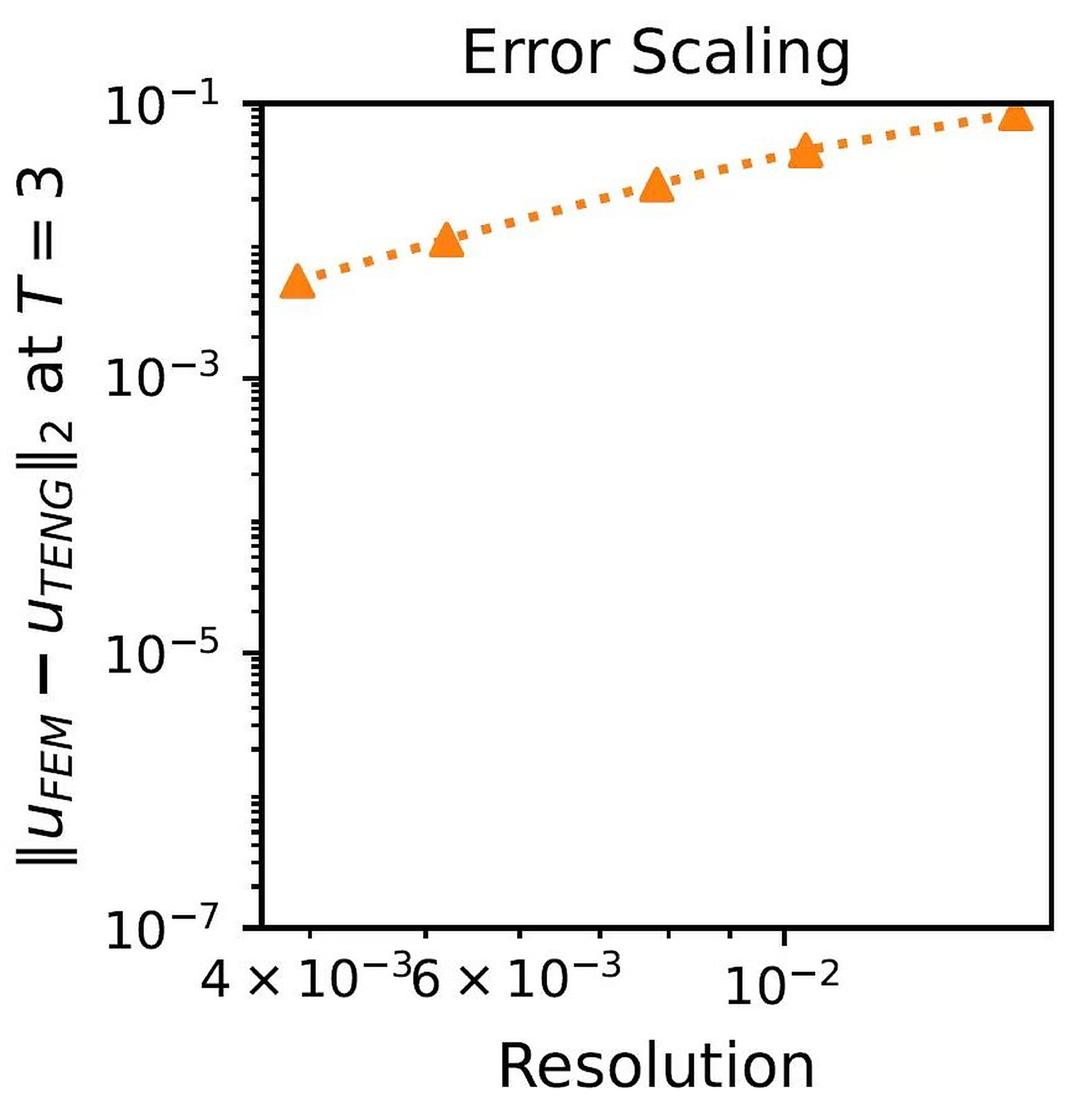}
        \caption{}
        \label{fig4-c:sub3}
    \end{subfigure}
    \begin{subfigure}{0.36\linewidth}
        \centering
        \includegraphics[width=\linewidth]{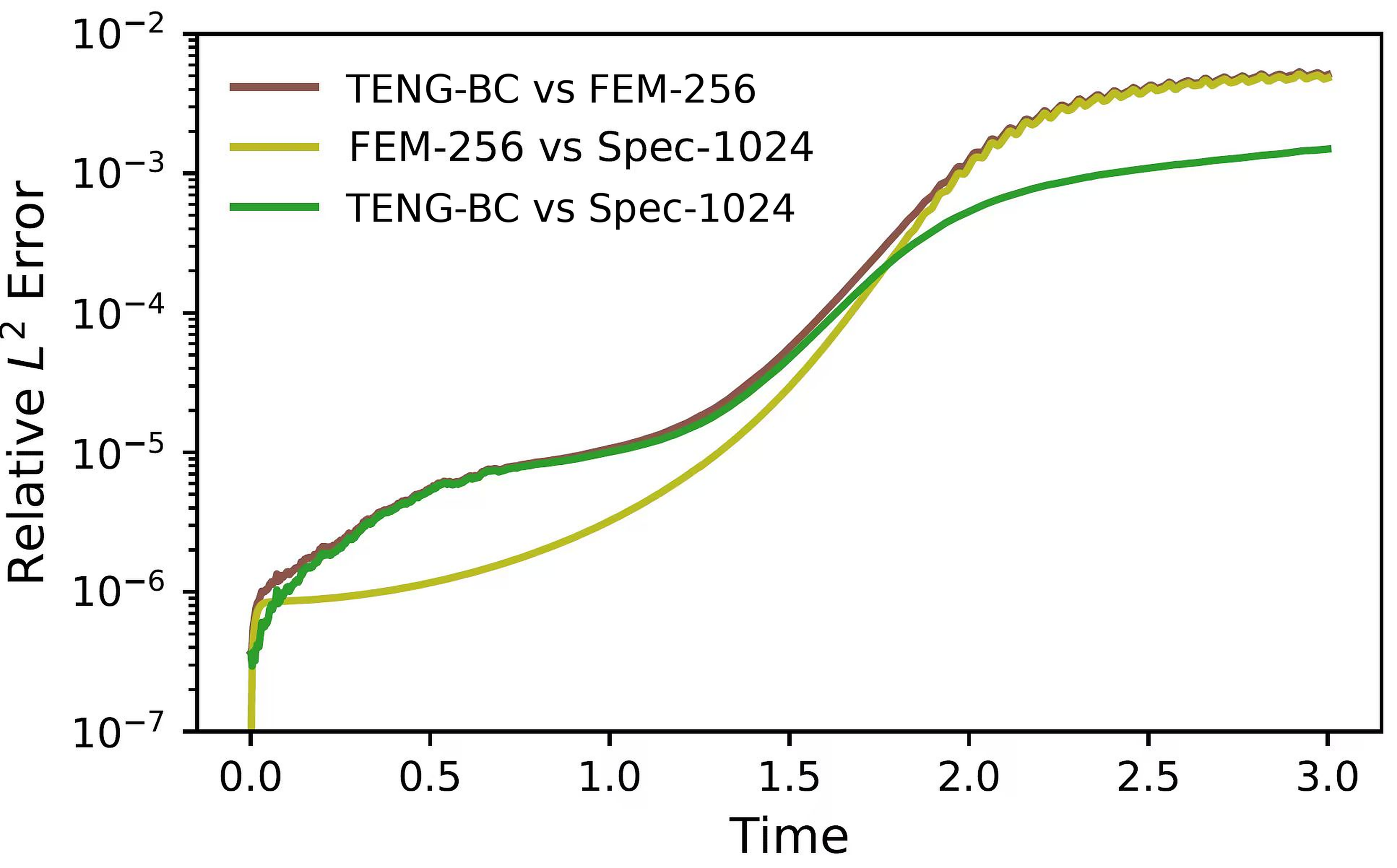}
        \caption{}
        \label{fig4-d:sub4}
    \end{subfigure}
    \caption{
    \textbf{Performance of TENG-BC on the Burgers equation with Dirichlet boundary.}
    \textbf{(a)} Predicted solutions by TENG-Heun at representative time steps.
    \textbf{(b)} Discrepancies between TENG-Heun and FEM solutions across multiple spatial resolutions. 
    \textbf{(c)} the scaling trend of discrepancies between TENG-Heun and FEM solutions at $T = 3$.
    \textbf{(d)} Pairwise $L^2$ discrepancies among TENG-Heun, FEM-256, and spectral-1024 solvers.
    }
    \label{fig4:burgers1}
    \vspace{-5pt}
\end{figure}

In our experiment, TENG-BC accurately reconstructs the nonlinear propagation and sharp front formation throughout the evolution on the circular domain $\{r\leq 1\}$ with Dirichlet boundary.
As illustrated in Figure~\ref{fig4-a:sub1}, the predicted fields by TENG-Heun remain sharply resolved yet physically smooth, showing no spurious oscillations or excessive numerical diffusion even near the discontinuity.
Residual errors are tightly localized along the shock front, while the remaining regions maintain near-perfect agreement with the spectral reference.

Figures~\ref{fig4-b:sub2}, \ref{fig4-c:sub3} and \ref{fig4-d:sub4} provide a detailed comparison between TENG-BC and FEM for the Burgers equation with Dirichlet boundary.
Figure~\ref{fig4-b:sub2} quantifies the $L^2$ discrepancy $\|u_{\text{FEM}}-u_{\text{TENG}}\|_2$ between TENG-Heun and FEM solutions across spatial resolutions and time steps.
As the FEM resolution increases, this difference decreases monotonically, indicating that FEM progressively converges toward the TENG-BC solution.
Figure~\ref{fig4-c:sub3} further shows the error scaling at $T=3$ with respect to FEM resolution, exhibiting a smooth, nearly linear decrease in log--log scale.
This trend persists up to FEM-256, beyond which refinement gives diminishing improvements, suggesting that TENG-BC’s continuous functional representation exceeds high-resolution FEM accuracy.

As the Burgers dynamics evolve, the three solvers begin to diverge in distinct ways.
Pairwise comparisons among TENG-Heun, FEM-256, and spectral-1024 in Figure~\ref{fig4-d:sub4} reveal that the FEM-related differences rise abruptly around $T\!\approx\!1.5$, coinciding with the emergence of the steep shock.
In contrast, the TENG–Spectral curve exhibits a much milder and delayed increase, implying that both methods continue to represent the evolving front coherently until substantially stronger nonlinearities develop.

\begin{table}[t]
    \caption{
        \textbf{Relative $L^2$ errors of different solvers for the Burgers equation with Dirichlet boundary conditions, measured against the spectral-1024 reference.}
        }\label{tab:burgers}%
    \vspace{5pt}
    \centering
    \begin{tabular}{@{}llll@{}}
    \toprule
    $L^2$-error$\downarrow$ & $T=1$  & $T=2$ & $T=3$ \\
    \midrule
    TENG-Euler & 3.29e-04 & 1.99e-03 & 3.02e-03  \\
    TENG-Heun & 9.98e-06 & \textbf{5.37e-04} & \textbf{1.49e-03}  \\
    FEM-128 & 4.01e-05 & 6.66e-03 & 2.58e-02  \\
    FEM-256 & \textbf{3.22e-06} & 1.14e-03 & 4.84e-03  \\
    PINN-BFGS & 1.28e-01 & 1.99e-01 & 2.08e-01 \\
    PINN-ENGD & 1.45e-01 & 2.33e-01 & 2.75e-01
    \\
    CoPINN & 2.42e-02 & 4.23e-02 & 4.89e-02
    \\
    \bottomrule
    \end{tabular}
    \vspace{-10pt}
\end{table}

A quantitative comparison against the spectral-1024 reference is summarized in Table~\ref{tab:burgers}.
At $T=1$, FEM-256 achieves the lowest error, benefiting from its high spatial resolution.
Notably, FEM-256 employs $513^2$ degrees of freedom, substantially more than the $65{,}536$ spatial sampling points used by TENG-BC.
In comparison, FEM-128 has $257^2$ degrees of freedom, which is comparable in number to the sampling points used by TENG-BC.
TENG-Heun consistently outperforms FEM-128 at all reported time points, while also surpassing the higher-resolution FEM-256 at $T=2$ and $T=3$.
These results highlight the competitive accuracy of TENG-BC, particularly as sharp solution gradients develop during the nonlinear evolution.

Together, these results highlight TENG-BC’s robustness: while all solvers encounter growing numerical challenges near the shock, TENG-BC maintains stable optimization and consistent solution quality well beyond the point where traditional solver begins to fail.

\section{Discussion}

We introduce TENG-BC to incorporate boundary constraints directly into the local optimization operator governing time evolution, without relying on penalty terms or boundary-specific trial-space constructions.
This unified framework accommodates Dirichlet, Neumann, Robin, and mixed boundary conditions, including heterogeneous and time-dependent constraints.
By avoiding loss-weight sensitivity and preserving coherent boundary evolution, TENG-BC achieves solver-level accuracy that matches or outperforms classical numerical solvers and PINN baselines across diverse PDE regimes.
By combining general boundary enforcement with high-accuracy time evolution, TENG-BC provides a practical framework for neural PDE solvers where boundary fidelity is essential.

\paragraph{Limitations and future work.}
Like other existing time-marching frameworks, TENG-BC remains subject to the intrinsic instability of the underlying dynamics. 
In highly unstable or chaotic regimes, accurate local least-squares projections cannot eliminate sensitivity to perturbations, and minute projection errors may compound over long time horizons. 
Extending TENG-BC to large-scale three-dimensional problems may incur substantial computational costs, while complex geometries may introduce errors through approximated boundary normals.
Promising directions for future work include extending the boundary-aware geometry to Cauchy-type constraints and broader classes of dynamical boundary value problems, as well as devising adaptive update strategies that further mitigate long-horizon error accumulation.

\section*{Impact Statement}

This work develops a neural numerical method for solving time-dependent PDEs with general boundary conditions.
Its primary potential impact lies in improving scientific computing tools for modeling physical systems in areas such as fluid dynamics, heat transfer, and engineering simulation.
By enabling accurate boundary enforcement in neural PDE solvers, the proposed framework may contribute to more flexible and efficient computational modeling workflows.

The work is methodological and does not involve human subjects, personal data, or direct deployment in safety-critical systems.
Potential negative impacts are therefore indirect and mainly concern the general risks of numerical simulation tools, such as misuse in applications where inaccurate physical predictions could inform high-stakes decisions.
Careful validation against trusted reference solvers remains necessary before applying such methods in safety-critical scientific or engineering settings.

\section*{Acknowledgement}
DL acknowledges support from the Beijing Major Science and Technology Project under Contract No. Z251100008125036.



\bibliographystyle{IEEEtran}
\bibliography{reference.bib}


\newpage
\appendix
\onecolumn

\section{Derivation of the Boundary-Aware Least-Squares Formulation}
\label{appendix-a: ls drivation}
For completeness, we include the derivation of the boundary-aware least-squares stepper that was omitted from the main text.  
Starting from the boundary functional
\begin{equation*}
    \mathcal{L}_b(u_\theta) = \frac{1}{2}||a u_\theta + b \partial_n u_\theta - v||_{L^2(\partial\Omega)}^2,
\end{equation*}
each boundary condition enters the optimization through its own contribution.  
Here, the existence of $\partial_n u_\theta$ makes the direct extension from simply fitting the target function to the boundary-constrained case non-trivial. We treat $u_\theta$ and $\partial_n u_\theta$ as jointly parameterized quantities.  
The least-squares update that minimizes $\mathcal{L}_b$ then takes the form
\begin{align*}
    \Delta\theta = \arg\min_{\Delta\theta} \bigg\|
        &(v - a u_\theta - b \partial_n u_\theta)(\cdot)
        - a(\cdot) \sum_j \frac{\partial u_\theta}{\partial \theta_j}(\cdot) \Delta\theta_j \\
        &- b(\cdot) \sum_k \frac{\partial (\partial_n u_\theta)}{\partial \theta_k}(\cdot) \Delta\theta_k
    \bigg\|_{L^2(\partial\Omega)}^2.
\end{align*}
This can be compactly written as
\begin{equation*}
    \Delta\theta = \arg\min_{\Delta\theta}
    \|\Delta v - (aJ + bK) \Delta\theta\|_{L^2(\partial\Omega)}^2,
\end{equation*}
where $J$, $K$ and $\Delta v$ are defined as in Equation~\ref{eq:boundary_ls}.  

When $(a,b,v)=(1,0,u_{\text{target}})$, the boundary-consistency term becomes identical to the interior consistency objective.
In this case, the interior and boundary contributions share the same residual structure and can be written in a unified form over the domain closure $\overline{\Omega}$:
\begin{equation}
\label{eq:full_ls_unified}
    \Delta\theta = \arg\min_{\Delta\theta}
    \|\Delta v - (aJ + bK) \Delta\theta\|_{L^2(\overline{\Omega})}^2,
\end{equation}
which is equivalent to the combined interior--boundary formulation in Equation~\ref{eq:full_ls} from the main text.
The function norm $L^2(\overline{\Omega})$ denotes the combined interior–boundary $L^2$ norm, incorporating both $\int_\Omega$ and $\int_{\partial\Omega}$ contributions.
This expression defines the boundary-aware least-squares stepper used throughout the algorithm.

\section{Theoretical Interpret as Natural Gradient}
\label{appendix-b: natgrad}
We show that the TENG-BC stepper can be viewed as a natural-gradient flow in parameter space.  
Starting with the simple case where the boundary loss is not included, the temporal evolution of the PDE solution satisfies $u_t = Lu$, and the natural-gradient flow seeks a parameter update $\Delta\theta$ such that the induced function variation $\delta u = J(\theta)\Delta\theta$ best approximates this evolution in the $L^2$ sense:
\begin{align*}
    \Delta\theta &= \arg\min_{\Delta\theta}\| (u_\text{target} - u_\theta)- J(\theta)\Delta\theta\|_{L^2(\Omega)}^2\\
    &= \arg\min_{\Delta\theta}\|-\frac{\partial\mathcal{L}_e}{\partial u_\theta} - J(\theta)\Delta\theta\|_{L^2(\Omega)}^2
\end{align*}
whose normal equation gives
\begin{align*}
    G(\theta)\Delta\theta &= -\int_\Omega J^T \frac{\partial\mathcal{L}_e}{\partial u_\theta}\,\mathrm{d}x = -\nabla_\theta\mathcal{L}_e, \\
    G(\theta) &= \int_\Omega J^TJ\,\mathrm{d}x.
\end{align*}
This leads to the standard natural-gradient update
\begin{equation}
    \label{eq2:nat_grad_update}
    \Delta\theta = -G(\theta)^{-1}\nabla_\theta\mathcal{L}_e(u_\theta),
\end{equation}
where the metric tensor $G(\theta)$ takes the same form as the Hilbert gram matrix from \cite{muller2023achieving}, which defines the Riemannian geometry of the parameter manifold induced by the $L^2$ inner product in function space.

When boundary constraints are included, the update of $\theta$ can be described by Equation~\ref{eq:full_ls_unified}. For simplicity, we write the total loss $\mathcal{L}=\mathcal{L}_e+\mathcal{L}_b$ in the unified form of:
\begin{equation*}
    \mathcal{L} = \frac{1}{2}\|au_\theta+b\partial_nu_\theta-v\|_{L^2(\overline{\Omega})}^2.
\end{equation*}
It is easy to check that when $(a,b,v)=(1,0,u_\text{target})$, the interior integral term reduces to the interior residual. Noticing that $\frac{\partial\mathcal{L}}{\partial u_\theta}=-a\Delta v$ and $\frac{\partial\mathcal{L}}{\partial (\partial_n u_\theta)}=-b\Delta v$, the corresponding normal equation of Equation~\ref{eq:full_ls_unified} gives:
\begin{align*}
    G(\theta)\Delta\theta &= \int_{\overline\Omega} (aJ+bK)^T\Delta v \,\mathrm{d}x  = -\int_{\overline\Omega} J^T \frac{\partial\mathcal{L}}{\partial u_\theta}\,\mathrm{d}x -\int_{\overline\Omega} K^T \frac{\partial\mathcal{L}}{\partial (\partial_nu_\theta)}\,\mathrm{d}x = -\nabla_\theta\mathcal{L}, \\
    G(\theta) &= \int_{\overline\Omega} (aJ+bK)^T(aJ+bK)\,\mathrm{d}x.
\end{align*}
Therefore the same formulation in Equation~\ref{eq2:nat_grad_update} extends naturally to the closure $\overline{\Omega}$ by augmenting the metric as
\begin{equation}
    \label{eq3:extended_gram_matrix}
    G(\theta) = \int_\Omega J^T J\,\mathrm{d}x + \int_{\partial\Omega} (aJ + bK)^T(aJ + bK)\,\mathrm{d}x,
\end{equation}
ensuring that both interior dynamics and boundary evolution are geometrically consistent within the same optimization operator. This expression shows that the inclusion of boundary terms simply extends the inner product that defines the Riemannian metric on the parameter manifold—from one defined purely over $\Omega$ to one over the closure $\overline{\Omega}$ that couples interior and boundary contributions.

Therefore, the boundary-aware least-squares stepper can be interpreted as a realization of a natural-gradient update with the parameter metric defined in Equation~\ref{eq3:extended_gram_matrix}.
This establishes the formal equivalence between the TENG-BC stepper and the natural-gradient flow, even in the presence of mixed boundary conditions.

\section{Error Analysis of the TENG-BC Stepper}
\label{appendix-c: error_bound}
\subsection{One-Step Error Decomposition}

Let $u(t)$ denote the exact solution and $u_\theta(t)$ the neural approximation at time $t$.
Consider one time step of size $\Delta t$.

For simplicity, we illustrate the analysis using the Euler time-stepping scheme.
The exact solution satisfies
\[
u(t+\Delta t)
=
u(t) + \Delta t\, L(u(t)) + \mathcal{O}(\Delta t^2).
\]

In TENG-BC, the target function is constructed as
\[
u_{\text{target}}
=
u_\theta(t) + \Delta t\, L(u_\theta(t)).
\]

The parameter update $\Delta\theta$ is obtained by solving a least-squares problem that matches the linearized neural output to $u_{\text{target}}$.

Therefore, the one-step error can be decomposed as
\[
u(t+\Delta t) - u_{\theta+\Delta\theta}
=
\underbrace{u(t+\Delta t) - u_{\text{target}}}_{\text{time discretization error}}
+
\underbrace{u_{\text{target}} - u_{\theta+\Delta\theta}}_{\text{projection error}}.
\]

\subsection{Error Propagation}

Let $e_n = u(t_n) - u_\theta(t_n)$ denote the error at step $n$.

Due to the dependence of the update on the current approximation, perturbations propagate across steps.
Assuming a stability factor $G$ associated with the underlying dynamics, which characterizes how perturbations are amplified across time steps, the error satisfies
\[
e_{n+1}
\;\le\;
G e_n + \tau_n + r_n.
\]

Unrolling the recursion gives
\[
e_n
\;\le\;
G^n e_0
+
\sum_{k=0}^{n-1} G^{\,n-k} (\tau_k + r_k).
\]

When $G \le 1$, the error remains controlled and is dominated by the accumulation of time discretization and projection residuals. This establishes the error bound stated in the main text.

\subsection{Discussion}

This analysis shows that the total error is governed by the accumulation of time discretization and projection residuals.
In particular, the least-squares formulation directly minimizes the projection residual $r_n$ at each step, leading to improved control of error propagation.

\section{Experimental Settings}
\label{appendix-c: settings}
\subsection{Neural Network Architectures and Optimization Details}

\paragraph{Network architecture.} Specifically, we adopt a multilayer perceptron (MLP) with $7$ hidden layers and width $40$ per layer, using activation function $tanh(\cdot)$ to ensure sufficient regularity of the neural representation.
The network parameters at time step $t$ are denoted by $\theta^{(t)}$, and the temporal evolution of the solution is realized through a sequence of parameter updates $\theta^{(t)} \rightarrow \theta^{(t+1)}$, rather than by introducing time as an explicit input variable.

\paragraph{Sampling.}
In all experiments, the collocation points used for the local optimization are fixed across time steps and are placed on a uniform grid over the spatial domain.
Both interior and boundary points remain identical throughout the temporal evolution, and no resampling is performed between successive optimization steps.
We empirically observe that alternative sampling strategies, including resampling at each time step or using randomly distributed points, do not lead to noticeable differences in the final solution accuracy or convergence behavior.
Therefore, a fixed grid-based sampling strategy is adopted for simplicity and reproducibility.

\paragraph{Time evolving schemes}
We use the same time evolving schemes as in \cite{chen2024teng}. The update of Heun scheme can be write as:
\begin{align*}
    k_1 &= Lu^{(t)} \\
    u^{(t+1/2\Delta t)} &= u^{(t)} + k_1\Delta t \\
    k_2 &= Lu^{(t+1/2\Delta t)} \\
    u_\text{target} &= u^{(t)} + \frac{1}{2}(k_1 + k_2)\Delta t.
\end{align*}
The update of RK4 scheme can be write as:
\begin{align*}
    k_1 &= Lu^{(t)} \\
    u^{(t+1/4\Delta t)} &= u^{(t)} + \frac{1}{2}k_1\Delta t \\
    k_2 &= Lu^{(t+1/4\Delta t)} \\
    u^{(t+1/2\Delta t)} &= u^{(t)} + \frac{1}{2}k_2\Delta t \\
    k_3 &= Lu^{(t+1/2\Delta t)} \\
    u^{(t+3/4\Delta t)} &= u^{(t)} + k_3\Delta t \\
    k_4 &= Lu^{(t+3/4\Delta t)} \\
    u_\text{target} &= u^{(t)} + \frac{1}{6}(k_1 + 2k_2 + 2k_3 + k_4)\Delta t.
\end{align*}

\paragraph{Optimization hyperparameters.}
At each time step, the local least-squares problem is solved using a fixed number of optimization iterations.
Unless otherwise specified, we perform $5$ optimization steps per time level.
To control computational cost and improve numerical stability, at each time step, $512$ parameters are allowed to vary, while the remaining parameters are held fixed.
The subset of active parameters is randomly chosen at each optimization step.

For multi-stage time discretization schemes such as Heun's method and the fourth-order Runge--Kutta (RK4) method, the first stage involves a more accurate initialization of the intermediate target.
Accordingly, we perform $7$ optimization steps for the first target in these schemes and allow a larger active parameter subset of size $768$ at the first optimization step, while all subsequent stages follow the standard setting described above.
This strategy was found to provide stable and consistent convergence across all experiments.

\paragraph{Computational environment and other implementation details.}
All experiments are conducted on a single NVIDIA V100 GPU with 16GB memory.
Double-precision arithmetic is used throughout to ensure numerical stability.
The same random seed is used across different methods for data sampling and parameter initialization to ensure reproducibility.

\subsection{Heat Equation}
We consider a collection of heat equation experiments with different boundary condition configurations.
All experiments solve the heat equation
\begin{equation*}
    \partial_t u(x,t) = \kappa \Delta u(x,t), \quad x\in\Omega\subset\mathbb{R}^2
\end{equation*}
where $\kappa = 0.1$ is the diffusion coefficient.
Let $(r,\theta)$ be the polar coordinates. All simulations are conducted on the circular domain $\Omega = \{r\leq 1\}$, except for the mixed-boundary configuration, which is computed on a quarter-annulus region $\Omega = \{0.5\leq r\leq1,\,0\leq\theta\leq\frac{\pi}{2}\}$.

\paragraph{Homogeneous Dirichlet boundary conditions.}
We impose homogeneous Dirichlet boundary conditions,
\begin{equation*}
    u(x,t) = 0, \quad x \in \partial\Omega.
\end{equation*}
An analytic reference solution is constructed using the eigenfunctions of the Dirichlet Laplacian on the disk.
Specifically, for integers $m\ge 0$ and $n\ge 1$, let $j_{m,n}$ denote the $n$-th positive zero of the Bessel function $J_m$.
The disk harmonics are defined as
\begin{equation*}
    \psi_{m,n}(r,\theta, t)
    = J_m(j_{m,n} r)\cos(m\theta)\exp\left(-\kappa\, j_{m,n}^2 t\right).
\end{equation*}
In our experiments, the reference solution is given by the following finite-mode combination:
\begin{align*}
u_{\mathrm{ref}}(r,\theta,t)
= \frac{1}{4}\Big(
&\psi_{0,1}(r,\theta,t) - \tfrac{1}{4}\psi_{0,2}(r,\theta,t)
+ \tfrac{1}{16}\psi_{0,3}(r,\theta,t) - \tfrac{1}{64}\psi_{0,4}(r,\theta,t) \\
&+ \psi_{1,1}(r,\theta,t) - \tfrac{1}{2}\psi_{1,2}(r,\theta,t)
+ \tfrac{1}{4}\psi_{1,3}(r,\theta,t) - \tfrac{1}{8}\psi_{1,4}(r,\theta,t) \\
&+ \psi_{2,1}(r,\theta,t) + \psi_{3,1}(r,\theta,t) + \psi_{4,1}(r,\theta,t)
\Big).
\end{align*}
The initial condition is set as $u(x,0)=u_{\mathrm{ref}}(x,0)$, which is compatible with the homogeneous Dirichlet boundary.

\paragraph{Homogeneous Neumann boundary conditions.}
We impose homogeneous Neumann boundary conditions,
\begin{equation*}
    \partial_n u = 0, \quad x \in \partial\Omega.
\end{equation*}
An analytic reference solution is constructed using eigenfunctions of the Neumann Laplacian on the disk.
Let $j'_{m,n}$ denote the $n$-th positive zero of the derivative of the Bessel function $J_m'(z)$.
The corresponding real-valued Neumann disk harmonics are
\begin{equation*}
    \psi_{m,n}(r,\theta,t)
    = J_m(j'_{m,n} r)\cos(m\theta)\exp\left(-\kappa (j'_{m,n})^2 t\right).
\end{equation*}
In our experiments, the reference solution is given by the finite-mode combination
\begin{align*}
u_{\mathrm{ref}}(r,\theta,t)
= \frac{1}{4}\Big(
&\psi_{0,1}(r,\theta,t) + \tfrac{1}{4}\psi_{0,2}(r,\theta,t)
+ \tfrac{1}{16}\psi_{0,3}(r,\theta,t) + \tfrac{1}{64}\psi_{0,4}(r,\theta,t) \\
&- \psi_{1,1}(r,\theta,t) - \tfrac{1}{2}\psi_{1,2}(r,\theta,t)
- \tfrac{1}{4}\psi_{1,3}(r,\theta,t) - \tfrac{1}{8}\psi_{1,4}(r,\theta,t) \\
&+ \psi_{2,1}(r,\theta,t) + \psi_{3,1}(r,\theta,t) + \psi_{4,1}(r,\theta,t)
\Big).
\end{align*}
The initial condition is set as $u(x,0)=u_{\mathrm{ref}}(x,0)$, ensuring consistency with the Neumann boundary condition.

\paragraph{Inhomogeneous Neumann boundary conditions.}
To construct an inhomogeneous Neumann boundary setting, we augment the above reference solution of homogeneous Neumann condition with an additional term
\begin{equation*}
    \tilde{\psi}(r,\theta,t)=J_1(\lambda_0 r)\cos(\theta)\exp\left(-\kappa\lambda_0^2 t\right),
\end{equation*}
where $\lambda_0=j_{1,1}$ is chosen such that $\tilde{\psi}$ does not satisfy the homogeneous Neumann condition on $r=1$.
The resulting boundary flux is then induced by this term, i.e.,
\(
\partial_n u\big|_{r=1}
=
\partial_n\tilde{\psi}\big|_{r=1}
\).

\paragraph{Robin boundary conditions. }
We impose Robin boundary conditions,
\begin{equation*}
    u +  \lambda \partial_n u = 0, \quad x \in \partial\Omega.
\end{equation*}

An analytic reference solution is constructed from a single separable disk harmonic:
\begin{equation*}
    u_{\mathrm{ref}}(r,\theta,t)
    = J_2(4r)\cos(2\theta)\,\exp\!\left(-\kappa\cdot 4^2\,t\right).
\end{equation*}

The Robin coefficient $\lambda$ is chosen such that $u_{\mathrm{ref}}$ satisfies the boundary condition at $r=1$.
Imposing $\lambda \partial_n u_{\mathrm{ref}}+u_{\mathrm{ref}}=0$ yields
\begin{equation*}
    \lambda = -\frac{J_2(4)}{4J_2'(4)}
    = \frac{J_2(4)}{4\big(\tfrac12 J_2(4)-J_1(4)\big)},
\end{equation*}
where the final expression uses standard Bessel-function recurrence identities.
The initial condition is set as $u(x,0)=u_{\mathrm{ref}}(x,0)$.

\paragraph{Mixed boundary conditions. }
We consider the heat equation on a sectorial annulus
\[
\Omega = \{(r,\theta):r\in[1/2,1], \theta\in[0,\pi/2]\},
\]
which corresponds to a quarter annulus in polar coordinates.

Different types of boundary conditions are imposed on different parts of $\partial\Omega$:
\begin{itemize}
    \item \textbf{Radial boundaries} ($\theta=0$ and $\theta=\pi/2$): homogeneous Neumann boundary conditions,
    \[
        \partial_\theta u = 0,
    \]
    corresponding to zero flux across the radial edges.
    \item \textbf{Outer circular boundary} ($r=1$): homogeneous Dirichlet boundary condition,
    \[
        u = 0.
    \]
    \item \textbf{Inner circular boundary} ($r=1/2$): inhomogeneous Dirichlet boundary condition,
    \[
        u = g(\theta),
    \]
    where the boundary profile $g$ is induced by the reference solution described below.
\end{itemize}
This configuration yields a genuinely mixed boundary setting combining Neumann and inhomogeneous Dirichlet constraints on geometrically distinct boundary segments.

An analytic reference solution is constructed using a finite combination of disk harmonics, restricted to the sectorial annulus.
Let $j_{m,n}$ denote the $n$-th positive zero of the Bessel function $J_m$.
For integers $m\ge 0$ and $n\ge 1$, define
\begin{equation*}
    \psi_{m,n}(r,\theta,t)
    = J_m(j_{m,n} r)\cos(m\theta)\exp\left(-\kappa j_{m,n}^2 t\right).
\end{equation*}
The reference solution is given by the finite linear combination
\begin{align*}
u_{\mathrm{ref}}(r,\theta,t)
= \frac{1}{4}\Big(
&\psi_{0,1}(r,\theta,t) - \tfrac{1}{2}\psi_{0,2}(r,\theta,t) + \tfrac{1}{4}\psi_{0,3}(r,\theta,t) - \tfrac{1}{8}\psi_{0,4}(r,\theta,t) \\
&+ \psi_{2,1}(r,\theta,t) - \tfrac{1}{2}\psi_{2,2}(r,\theta,t) + \tfrac{1}{4}\psi_{2,3}(r,\theta,t) - \tfrac{1}{8}\psi_{2,4}(r,\theta,t) \\
&+ \psi_{4,2}(r,\theta,t)
\Big),
\end{align*}
where all modes are evaluated at $(r,\theta,t)$.
The initial condition is set as $u(x,0)=u_{\mathrm{ref}}(x,0)$.

\subsection{Transport Equation}
We consider a first-order transport equation on the unit disk
\[
\Omega=\{x=(x_1,x_2)\in\mathbb{R}^2: x_1^2+x_2^2\le 1\}.
\]
The scalar field $u(x,t)$ evolves according to
\begin{equation*}
    \partial_t u + v(x)\cdot \nabla u = 0, \quad x\in\Omega,
\end{equation*}
where the velocity field $v(x)$ is defined by
\begin{equation*}
    v(x_1,x_2)=\big(-\beta x_1 + x_2,-\beta x_2 - x_1\big),
\end{equation*}
so that the advection operator is $v\cdot \nabla u
= -(\beta x_1 - x_2)\partial_{x_1}u - (\beta x_2 + x_1)\partial_{x_2}u$. We choose $\beta=0.2$.

We prescribe a Dirichlet boundary condition on $\partial\Omega$ induced by the reference solution, i.e., $u(x,t)=u_{\mathrm{ref}}(x,t)$ for $x\in\partial\Omega$. The initial condition is set as $u(x,0)=u_{\mathrm{ref}}(x,0)$.

The analytic reference solution is constructed by tracing the initial and inflow data along characteristic trajectories.
Let $j_{m,n}$ denote the $n$-th positive zero of the Bessel function $J_m$.
Define the time-dependent disk harmonics
\begin{equation*}
    \phi_{m,n}(r,\theta,t)
    = J_m\big(j_{m,n} r e^{\beta t}\big)\cos\big(m(\theta+t)\big).
\end{equation*}
The reference solution is given by the finite linear combination
\begin{align*}
u_{\mathrm{ref}}(r,\theta,t)=\frac{1}{4}\Big(
&\phi_{0,1}(r,\theta,t)-\tfrac{1}{4}\phi_{0,2}(r,\theta,t)+\tfrac{1}{16}\phi_{0,3}(r,\theta,t)-\tfrac{1}{64}\phi_{0,4}(r,\theta,t)\\
&+\phi_{1,1}(r,\theta,t)-\tfrac{1}{2}\phi_{1,2}(r,\theta,t)+\tfrac{1}{4}\phi_{1,3}(r,\theta,t)-\tfrac{1}{8}\phi_{1,4}(r,\theta,t) \\
&+\phi_{2,1}(r,\theta,t)+\phi_{3,1}(r,\theta,t)+\phi_{4,1}(r,\theta,t)
\Big).
\end{align*}

\subsection{Burgers Equation}
We consider the viscous Burgers equation on the circular domain,
\begin{equation*}
    u_t + u(\partial_{x_1}u+\partial_{x_2}u) = \nu\nabla^2u, \quad x=(x_1, x_2)\in \Omega = \{r\leq 1\}.
\end{equation*}
with diffusion coefficient $\nu = 10^{-3}$.
The initial condition is defined as
\begin{align*}
u(x_1,x_2,0)
= \frac{1}{100}\Big(
&\exp\big(3\sin x_1 + \sin x_2\big)
+ \exp\big(-3\sin x_1 + \sin x_2\big) \\
&-\exp\big(3\sin x_1 - \sin x_2\big)
-\exp\big(-3\sin x_1 - \sin x_2\big)
\Big).
\end{align*}
To get accurate reference solution, we extend it to the square domain $[-\pi, \pi]^2$ with periodic boundary conditions along both axes:
\[
u(x_1+2\pi,x_2)=u(x_1,x_2),\qquad
u(x_1,x_2+2\pi)=u(x_1,x_2).
\]
And reference solution $u_{\mathrm{ref}}$ is given by spectral method. The Dirichlet boundary condition of the original problem on $\partial\Omega$ induced by the reference solution, i.e., $u(x,t)=u_{\mathrm{ref}}(x,t)$ for $x\in\partial\Omega$.

\subsection{Baseline Methods and Implementation Details}
\paragraph{Physics-Informed Neural Networks (PINNs).}
As a learning-based baseline, we adopt a standard Physics-Informed Neural Network (PINN) implementation following the setup in prior work \cite{muller2023achieving}.
The solution $u(x,t)$, represented by a neural network, is trained by minimizing a global space--time residual loss, with initial and boundary conditions enforced via additional penalty terms.

Optimization is performed using both BFGS and ENGD algorithm, and each experiment is trained for $100{,}000$ optimization steps.
Since \cite{muller2023achieving} only provide the implementation for heat equation with dirichlet boundaries, we omit the benchmark of ENGD optimizer for other cases.
For all problems, global collocation points are sampled uniformly, with $100{,}000$ interior points, $15{,}000$ initial-condition points, and $10{,}000$ boundary points.
The trained PINN solutions are evaluated on the same grids and compared against the same reference solutions using identical error metrics.

\paragraph{Cognitive Physics-Informed Neural Networks (CoPINN).}
As a recent PINN-based baseline, we additionally consider CoPINN~\cite{duancopinn}, which improves training efficiency through cognitive sampling and curriculum-based optimization.
Similar to standard PINNs, the neural solution is trained by minimizing a global space--time residual objective with boundary and initial conditions enforced through penalty terms.

For fair comparison, CoPINN uses the same network architectures and evaluation metrics as TENG++.
For all problems, global collocation points are sampled on a $256\times256\times256$ grid.
All experiments are trained for $50{,}000$ optimization steps and evaluated against the same reference solutions on identical grids.

\paragraph{Finite Element Method (FEM).}
As a classical numerical baseline, we implement the finite element method (FEM) based on \cite{skfem2020}.
For problems defined on rectangular domains (e.g., the Burgers equation), we employ structured square meshes with $N^2$ grid points, where $N$ denotes the spatial resolution.
For circular and other non-rectangular domains, unstructured triangular meshes are used, with an effective spatial resolution set to $256$ by default.
In all cases, second-order finite elements are adopted for spatial discretization, and second-order time integration schemes are used for temporal discretization.
The time step size is fixed to $\Delta t = 10^{-4}$ for all experiments.
Boundary and initial conditions are imposed using standard FEM formulations.
The FEM solutions are evaluated on the same grids and compared against the same reference solutions as the proposed method.

\paragraph{Spectral method for the Burgers equation.}
For the Burgers equation with periodic boundary conditions on $[0,2\pi]^2$, we additionally employ a high-accuracy Fourier spectral method \cite{canuto2007spectral} to serve as a numerical reference.
The solution is expanded in Fourier modes along both spatial directions, and spatial derivatives are computed in the spectral domain.
Time integration is performed using a RK4 time-stepping scheme with a fixed step size $\Delta t = 10^{-4}$.
We use a spectral resolution of $1024\times 1024$ Fourier modes, which is sufficient to resolve all relevant scales in the solution.
The resulting solution is used for comparison with neural and FEM-based methods under the same evaluation protocol.

\section{Additional Experimental Results}

\begin{figure}[!h]
    \centering
    \includegraphics[width=0.75\linewidth]{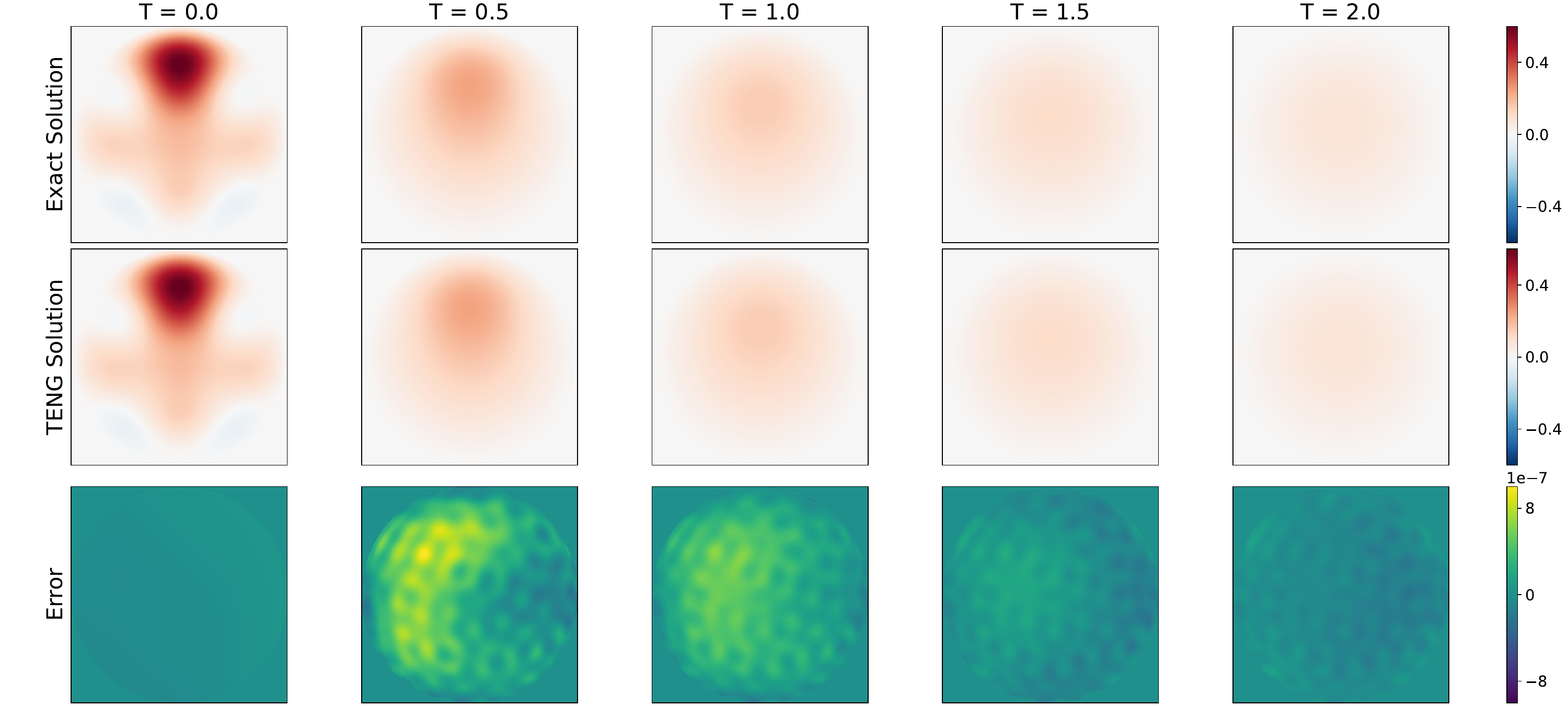}
    \caption{\textbf{Comparison between the exact solution and the TENG-BC prediction for the heat equation with dirichlet boundary at representative time instances.} Top: exact solution. Middle: solution obtained by TENG-BC. Bottom: absolute error.}
    \label{fig-appendix:heat_dirichlet}
\end{figure}

\begin{figure}[!h]
    \centering
    \includegraphics[width=0.75\linewidth]{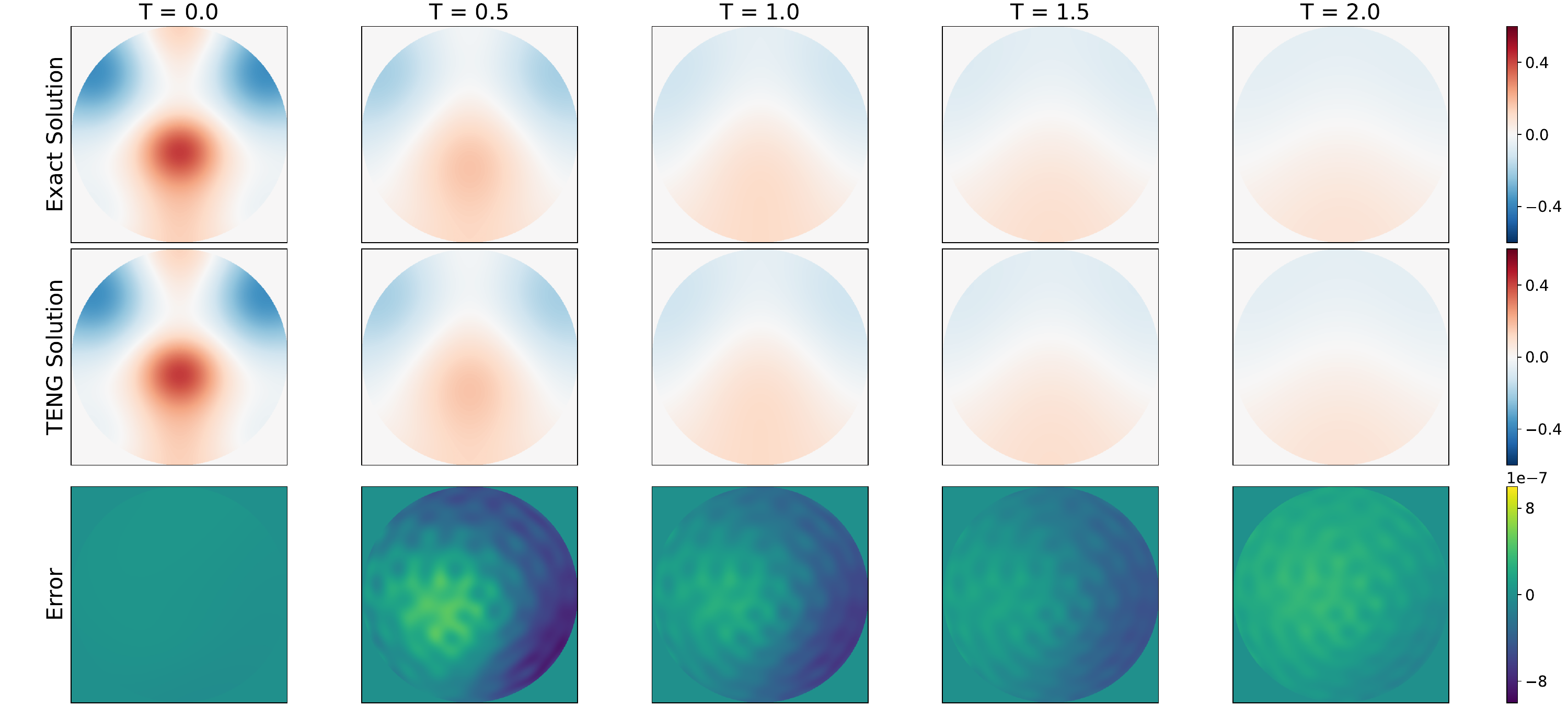}
    \caption{\textbf{Comparison between the exact solution and the TENG-BC prediction for the heat equation with zero neumann boundary at representative time instances.} Top: exact solution. Middle: solution obtained by TENG-BC. Bottom: absolute error.}
    \label{fig-appendix:heat_neumann}
\end{figure}

\begin{figure}[!h]
    \centering
    \includegraphics[width=0.75\linewidth]{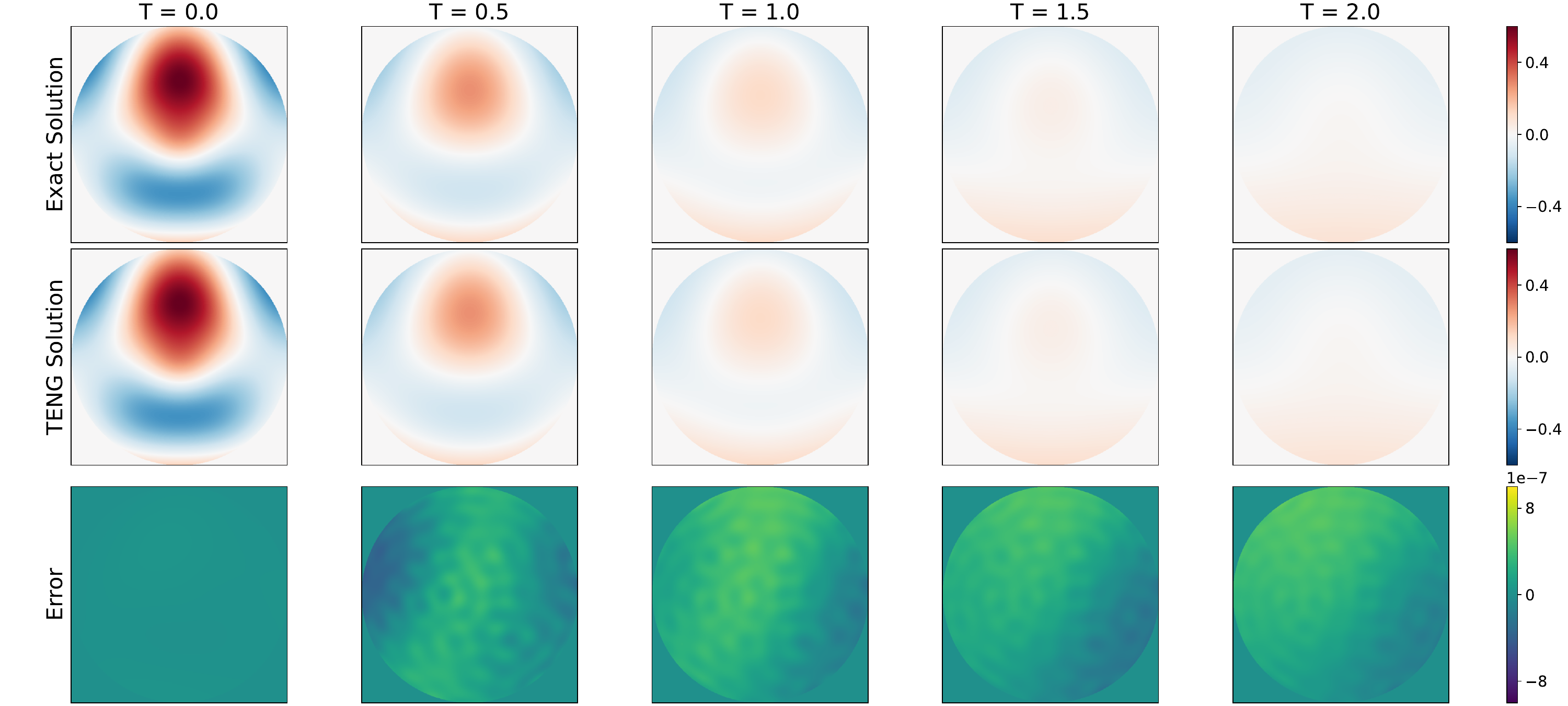}
    \caption{\textbf{Comparison between the exact solution and the TENG-BC prediction for the heat equation with nonzero neumann boundary at representative time instances.} Top: exact solution. Middle: solution obtained by TENG-BC. Bottom: absolute error.}
    \label{fig-appendix:heat_nonzero}
\end{figure}

\begin{figure}[!h]
    \centering
    \includegraphics[width=0.75\linewidth]{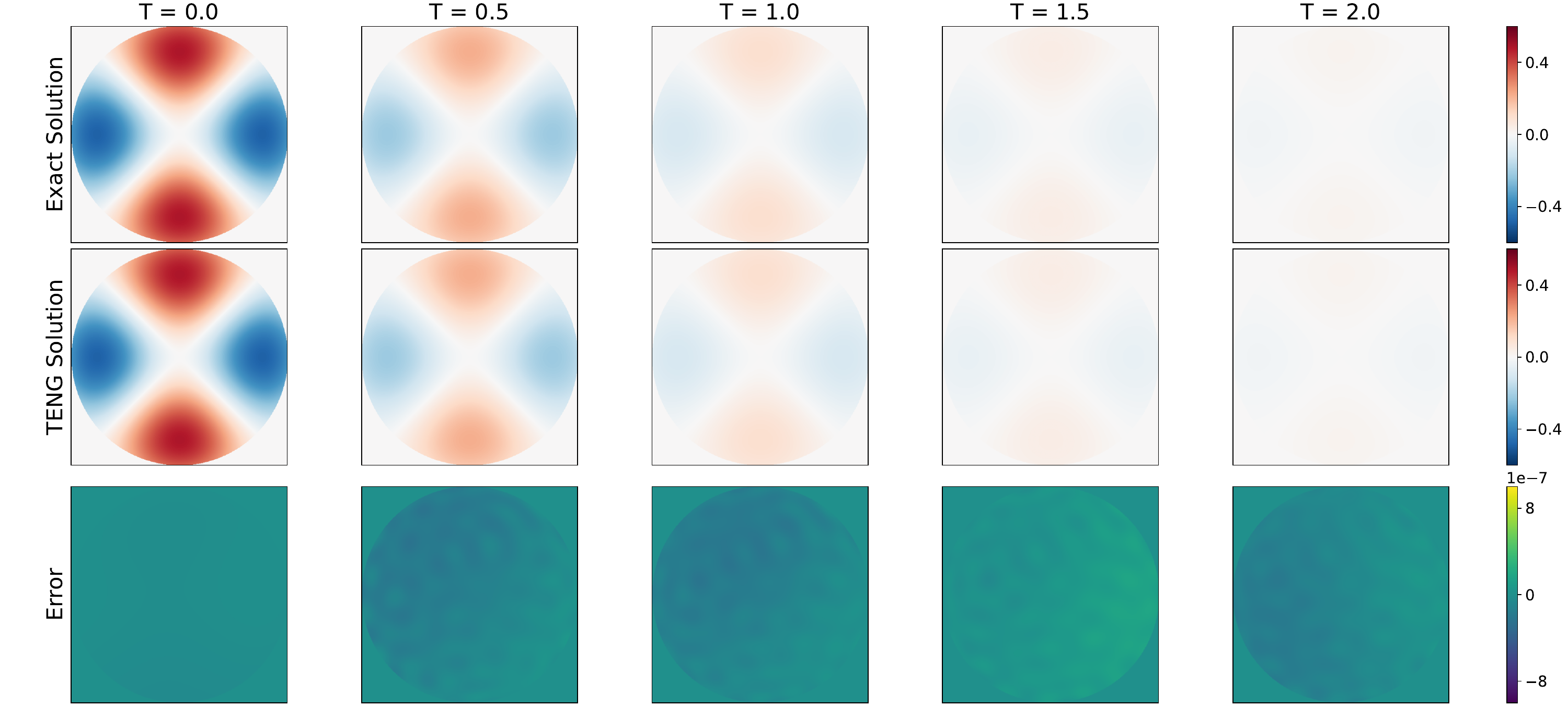}
    \caption{\textbf{Comparison between the exact solution and the TENG-BC prediction for the heat equation with robin boundary at representative time instances.} Top: exact solution. Middle: solution obtained by TENG-BC. Bottom: absolute error.}
    \label{fig-appendix:heat_robin}
\end{figure}

\begin{figure}[!h]
    \centering
    \includegraphics[width=0.75\linewidth]{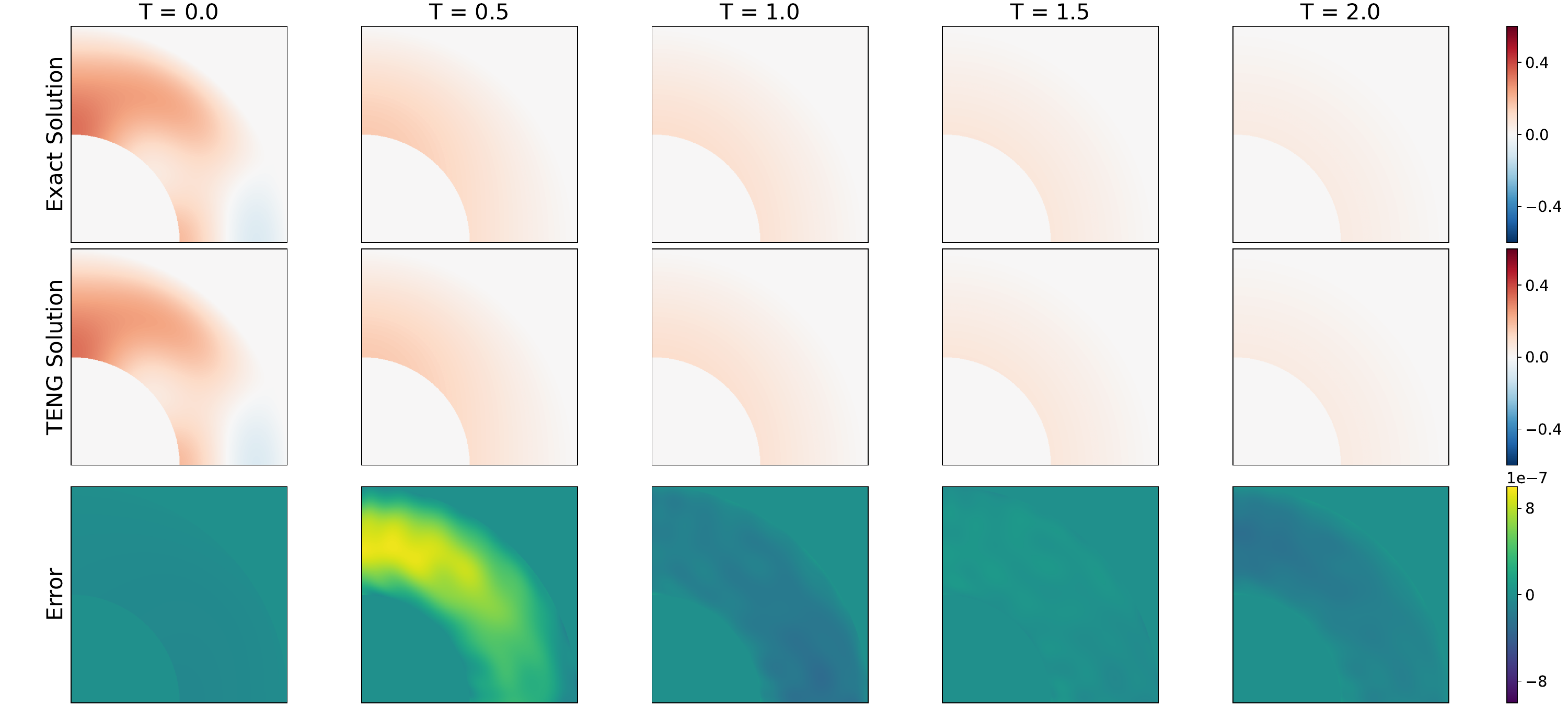}
    \caption{\textbf{Comparison between the exact solution and the TENG-BC prediction for the heat equation with mixed boundary at representative time instances.} Top: exact solution. Middle: solution obtained by TENG-BC. Bottom: absolute error.}
    \label{fig-appendix:heat_mix}
\end{figure}

\paragraph{Additional numerical visualization results for previous experiments. }
We first present additional qualitative comparisons between the exact solutions and the solutions obtained by TENG-BC for the heat equation.
All TENG-BC solutions shown here are computed using the TENG--Heun time discretization under the default experimental settings.
Representative solution snapshots are shown at selected time instances in Figure~\ref{fig-appendix:heat_dirichlet} to \ref{fig-appendix:heat_mix}.
The predicted fields produced by TENG-BC closely match the corresponding exact solutions in both global structure and fine-scale details.
No spurious oscillations or numerical artifacts are observed throughout the spatial domain.
The error remains uniformly small across the domain and does not concentrate near boundaries or regions with higher curvature.
This behavior indicates that the boundary-aware formulation in TENG-BC enforces the heat equation and its boundary conditions consistently at each time step, leading to stable and accurate long-term evolution.

\begin{figure}[!h]
    \centering
    \includegraphics[width=0.75\linewidth]{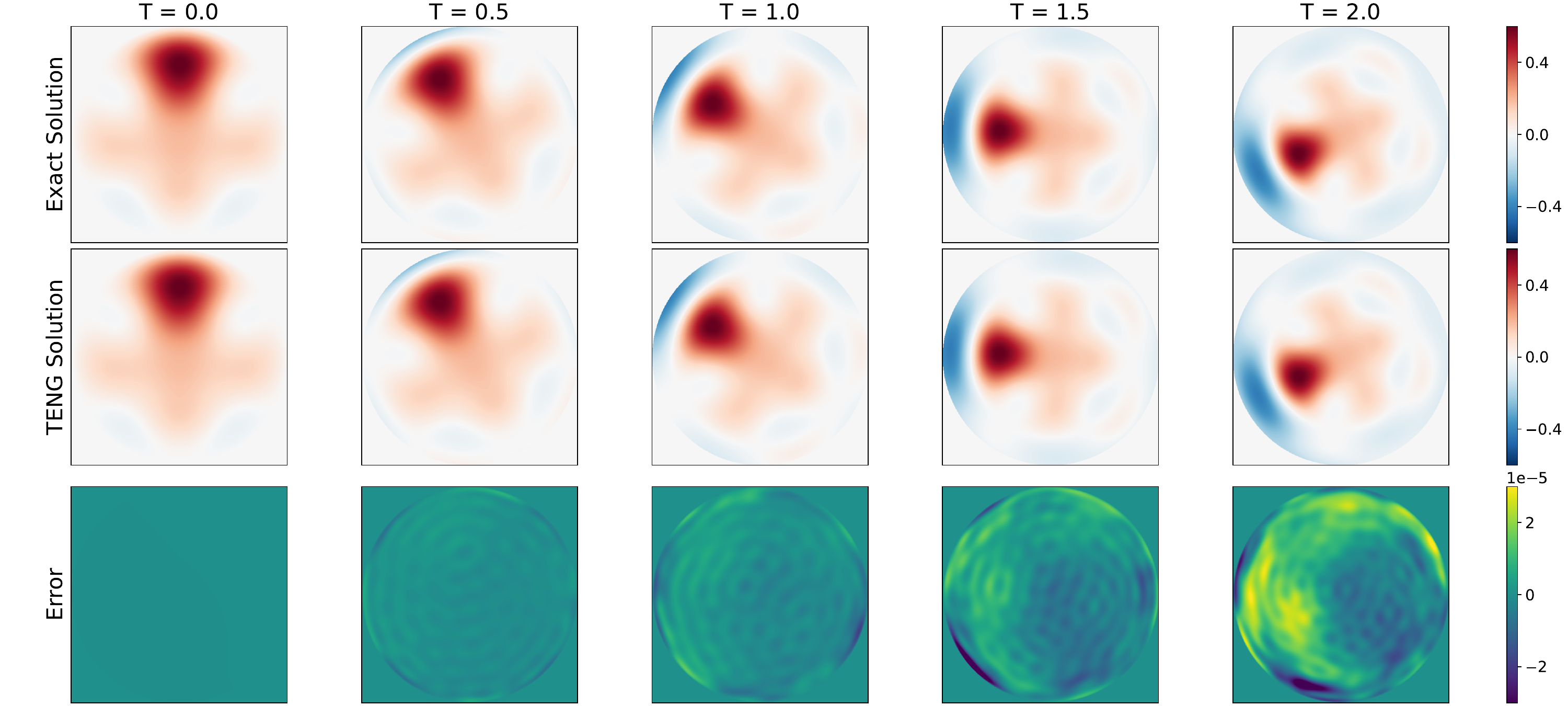}
    \caption{\textbf{Comparison between the exact solution and the TENG-BC prediction for the transport equation with dirichlet boundary at representative time instances.} Top: exact solution. Middle: solution obtained by TENG-BC. Bottom: absolute error.}
    \label{fig-appendix:transport}
\end{figure}

We next present the result of transport equation with dirichlet boundary, which poses a more challenging test due to its advection-dominated nature and the absence of intrinsic dissipation.
The TENG-BC solutions shown here are computed using the TENG--Heun time discretization under the default experimental settings.
Figure~\ref{fig-appendix:transport} compares the exact solution and the TENG-BC prediction at representative time instances.
The transported structures are accurately preserved, with both amplitude and phase faithfully reproduced by the neural solver.

\begin{figure}[!h]
    \centering
    \includegraphics[width=0.75\linewidth]{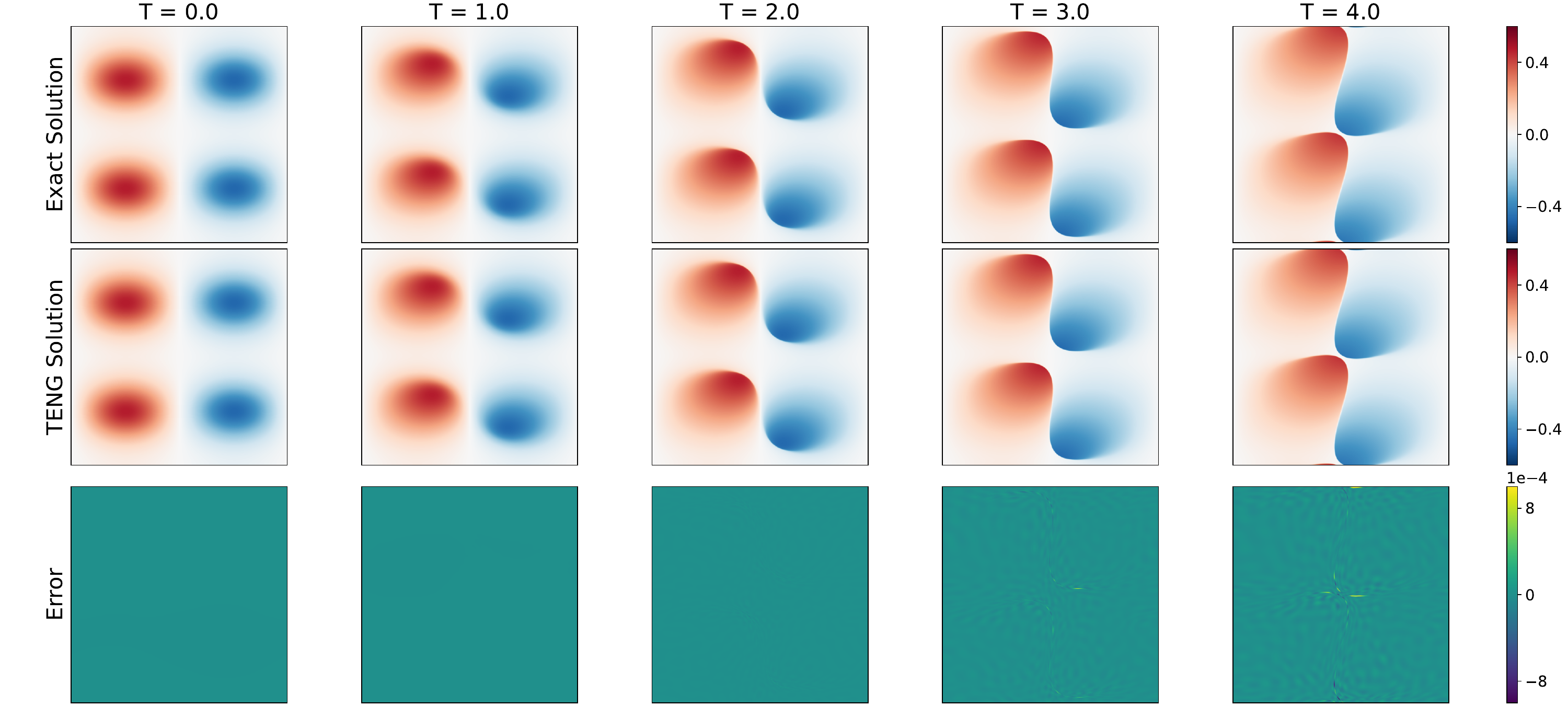}
    \caption{\textbf{Comparison between the exact solution and the TENG-BC prediction for the Burgers equation with periodic boundary at representative time instances.} Top: exact solution. Middle: solution obtained by TENG-BC. Bottom: absolute error.}
    \label{fig-appendix:burgers}
\end{figure}

We also include the result of Burgers equation with periodic boundary in Figure~\ref{fig-appendix:burgers}. 
The TENG-BC solutions shown here are computed using the TENG--Heun time discretization under the default experimental settings.
Because of the solution's shock structure, residual errors are tightly localized along the front, while the remaining regions maintain near-perfect agreement with the spectral reference. 


\begin{figure}[!h]
    \centering
    \includegraphics[width=0.95\linewidth]{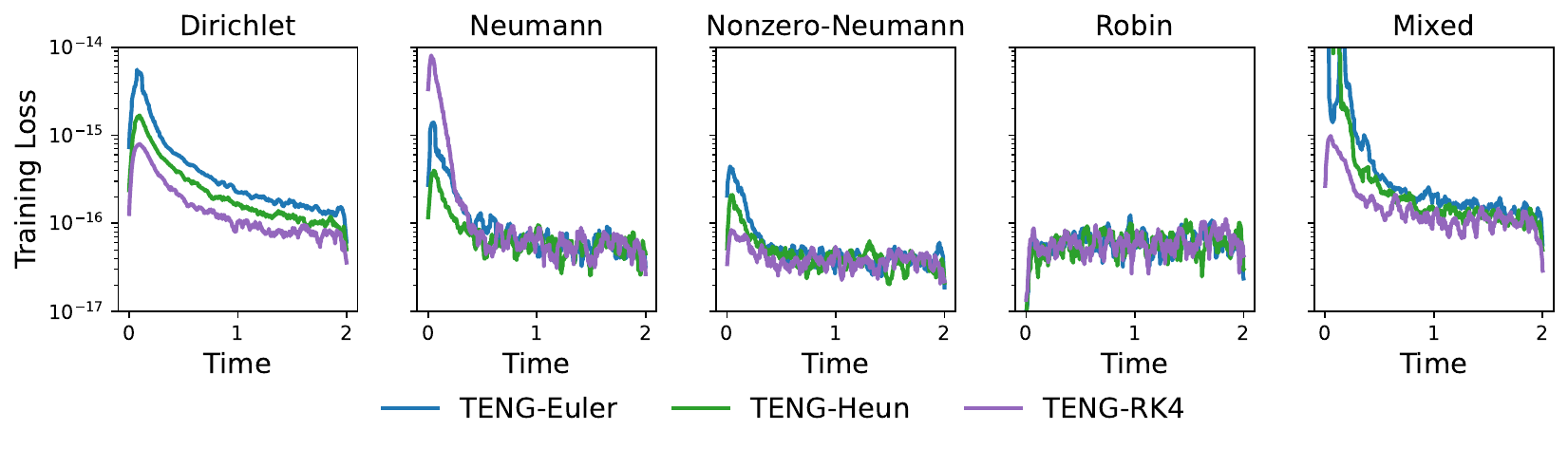}
    \caption{\textbf{Per-step loss of TENG-BC on the heat equation with different boundaries.} Here training loss means the per-step least-squares residuals showing local fitting accuracy within each update.}
    \label{fig-appendix:heat_loss}
\end{figure}
\begin{figure}[!h]
    \centering
    \includegraphics[width=0.5\linewidth]{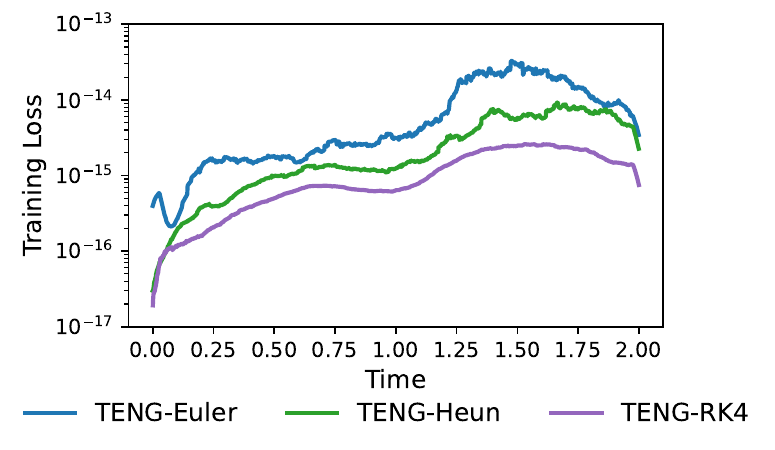}
    \caption{\textbf{Per-step loss of TENG-BC on the transport equation with dirichlet boundary.} Here training loss means the per-step least-squares residuals showing local fitting accuracy within each update.}
    \label{fig-appendix:transport_loss}
\end{figure}
\begin{figure}[!h]
    \centering
    \includegraphics[width=0.45\linewidth]{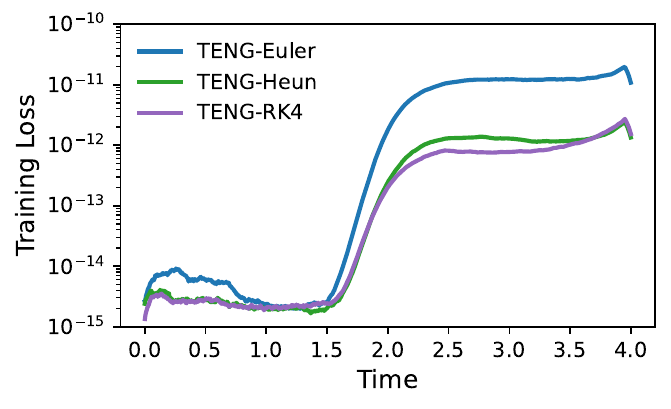}
    \caption{\textbf{Per-step loss of TENG-BC on the Burgers equation with Dirhclet boundary.} Here training loss means the per-step least-squares residuals showing local fitting accuracy within each update.}
    \label{fig-appendix:burgers_loss}
\end{figure}

\paragraph{Per-step optimization residuals.}
In Figure~\ref{fig-appendix:heat_loss}, \ref{fig-appendix:transport_loss} and \ref{fig-appendix:burgers_loss}, we report the per-step least-squares residuals of the TENG-BC updates across all benchmark problems.
These residuals measure the local fitting accuracy at each time step, reflecting how well the neural representation matches the operator-predicted target under the imposed boundary constraints. For heat and transport equations, in all experiments, the residuals remain stable and small throughout the entire temporal evolution, indicating that the boundary-aware least-squares stepper consistently achieves high-precision local updates independent of the underlying PDE type or boundary configuration.

For Burgers equation, The optimization process of TENG-BC reflects the same physical transition as shown by $L^2$-error in the main text.
In Figure~\ref{fig-appendix:burgers_loss}, the per-step fitting loss remains near machine precision for most of the evolution, then exhibits a small, well-localized rise around $T\!\approx\!1.8$.
This rise coincides with the physical onset of the shock and reflects the increased difficulty of fitting highly non-smooth target functions, rather than a failure of the optimization itself.
Importantly, the loss quickly stabilizes afterward, confirming that TENG-BC adapts to the new solution regime without divergence.

\paragraph{Sensitivity and robustness studies for Burgers Equation. }
We report additional sensitivity and robustness experiments for the Burgers equation.
These results are intended to assess the stability of the proposed method under variations in network architecture, optimization settings, and sampling strategies, rather than to optimize performance.
Unless otherwise stated, all experiments follow the same setup as in the main text.

\begin{figure}[htb]
    \centering
    \includegraphics[width=0.5\linewidth]{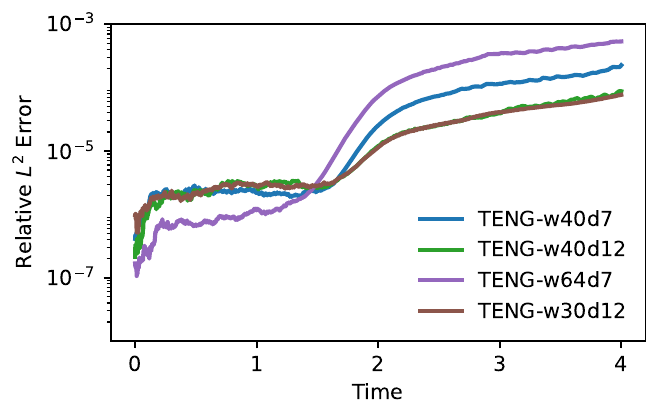}
    \caption{\textbf{Sensitivity of TENG to network size for the Burgers equation.} All experiments use MLP structure with different width and depth. The curve labeled TENG-w40d7 is the results by MLP with width 40 and depth 7 (the baseline architecture in previous experiments). The meanings of other labels are similar.}
    \label{fig-appendix:burgers1}
\end{figure}

We first examine the effect of network architecture on the performance of TENG for the Burgers equation.
Starting from the baseline architecture of width 40 and depth 7, we test wider and deeper variants, including width 40 depth 12, width 64 depth 7, and width 30 depth 12.
From the results in Figure~\ref{fig-appendix:burgers1}, we observe that increasing network depth leads to consistently improved approximation quality, while increasing width alone provides limited benefit.
Notably, the width 30 depth 12 architecture achieves performance comparable to width 40 depth 12 despite having a similar parameter count to the baseline, indicating that depth plays a more important role than width for this problem.

\begin{figure}[htb]
    \centering
    \includegraphics[width=0.48\linewidth]{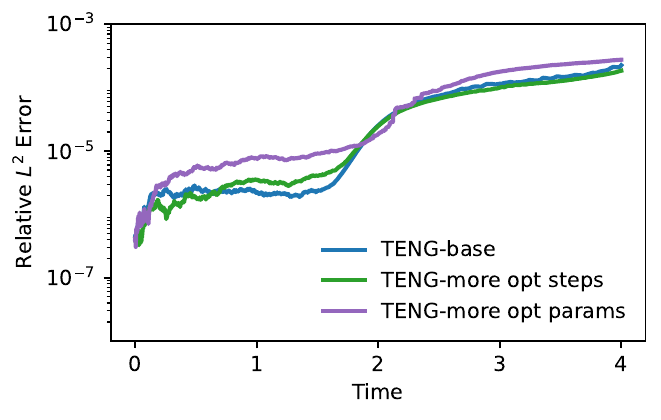}
    \includegraphics[width=0.48\linewidth]{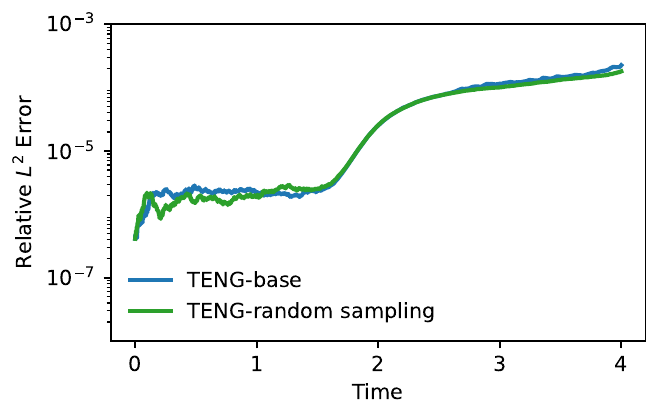}
    \caption{\textbf{Effect of per-step optimization settings and sampling strategy on the Burgers equation.} \textbf{Left:} different optimization strategies with more optimization steps (TENG--more opt steps) and more parameters updated per step (TENG--more opt params). \textbf{Right:} different sampling strategies with randomized sampling at each step (TENG--random sampling).}
    \label{fig-appendix:burgers2}
\end{figure}

We next study the robustness of TENG with respect to per-step optimization settings and sampling strategies.
Specifically, we vary the number of optimization steps performed at each time step and the number of trainable parameters updated per step.
The results are represented in Figure~\ref{fig-appendix:burgers2}.
Across all tested configurations, increasing either quantity does not lead to systematic performance improvements and in some cases slightly degrades accuracy. 
We also compare fixed grid sampling with per-step random sampling of spatial points.
The resulting solutions exhibit nearly identical accuracy and error evolution.
These results indicate that the proposed method is robust to both optimization hyperparameters and the choice of sampling strategy.

\paragraph{Sensitivity to Boundary Sampling Density.}
We investigate the effect of boundary sampling density on the heat equation with Dirichlet boundary conditions.
The number of interior samples is fixed at $N_\Omega=65{,}536$, while the number of sampling points per boundary edge varies from $16$ to $1024$.
As shown in Table~\ref{tab:boundary_sampling}, increasing the number of boundary samples generally improves boundary accuracy.
However, excessively dense boundary sampling may slightly increase the interior error, as the least-squares projection places greater emphasis on boundary constraints.
Overall, TENG-BC maintains high accuracy once sufficient boundary samples are included, demonstrating its robustness to the sampling configuration.

\begin{table}[t]
    \centering
    \caption{Sensitivity to boundary sampling density for the heat equation
    with Dirichlet boundary conditions.}
    \label{tab:boundary_sampling}
    \small
    \begin{tabular}{c|cc|cc}
        \toprule
        $N_{\partial\Omega}$ &
        \multicolumn{2}{c|}{$T=1$} &
        \multicolumn{2}{c}{$T=2$} \\
        & RMSE & bRMSE & RMSE & bRMSE \\
        \midrule
        16$\times$4 & 7.73e-04 & 2.64e-04 & 2.15e-03 & 5.17e-04 \\
        64$\times$4 & \textbf{3.39e-06} & 2.52e-07 & \textbf{3.06e-06} & 1.60e-07 \\
        256$\times$4 & 5.92e-06 & \textbf{4.76e-08} & 7.54e-06 & \textbf{2.81e-08} \\
        1024$\times$4 & 1.74e-05 & 1.83e-08 & 1.40e-05 & 1.15e-08 \\
        \bottomrule
    \end{tabular}
\end{table}



\end{document}